\documentclass{ieeeaccess}
\usepackage{cite}
\usepackage{amsmath,amssymb,amsfonts}
\usepackage{algorithmic}
\usepackage{graphicx}
\usepackage{multirow}
\usepackage{textcomp}
\usepackage{colortbl}
\usepackage{xcolor}
\usepackage{hhline}
\def\BibTeX{{\rm B\kern-.05em{\sc i\kern-.025em b}\kern-.08em
    T\kern-.1667em\lower.7ex\hbox{E}\kern-.125emX}}
\begin{document}
\history{Date of publication xxxx 00, 0000, date of current version xxxx 00, 0000.}
\doi{10.1109/ACCESS.2017.DOI}

\title{One-Shot Learning for Periocular Recognition: Exploring the Effect of Domain Adaptation and Data Bias on Deep Representations}
\author{\uppercase{Kevin Hernandez-Diaz}, \IEEEmembership{Student Member, IEEE},
\uppercase{Fernando Alonso-Fernandez}, \IEEEmembership{Member, IEEE} \uppercase{and Josef Bigun},
\IEEEmembership{Fellow, IEEE}}
\address[1]{Halmstad University, Halmstad, CO 30261 Sweden}


\markboth
{Hernandez-Diaz \headeretal: One-Shot Learning for Periocular Recognition}
{Hernandez-Diaz \headeretal: One-Shot Learning for Periocular Recognition}

\corresp{Corresponding author: Kevin Hernandez-Diaz (e-mail: kevin.hernandez-diaz@hh.se).}

\begin{abstract}
One weakness of machine-learning algorithms is the need to train the models for a new task. This presents a specific challenge for biometric recognition due to the dynamic nature of databases and, in some instances, the reliance on subject collaboration for data collection. In this paper, we investigate the behavior of deep representations in widely used CNN models under extreme data scarcity for One-Shot periocular recognition, a biometric recognition task. We analyze the outputs of CNN layers as identity-representing feature vectors. We examine the impact of Domain Adaptation on the network layers' output for unseen data and evaluate the method's robustness concerning data normalization and generalization of the best-performing layer. We improved state-of-the-art results that made use of networks trained with biometric datasets with millions of images and fine-tuned for the target periocular dataset by utilizing out-of-the-box CNNs trained for the ImageNet Recognition Challenge and standard computer vision algorithms. For example, for the Cross-Eyed dataset, we could reduce the EER by $67\%$ and $79\%$ (from 1.70$\%$ and 3.41$\%$ to 0.56$\%$ and 0.71$\%$) in the Close-World and Open-World protocols, respectively, for the periocular case. We also demonstrate that traditional algorithms like SIFT can outperform CNNs in situations with limited data or scenarios where the network has not been trained with the test classes like the Open-World mode. SIFT alone was able to reduce the EER by $64\%$ and $71.6\%$ (from 1.7$\%$ and 3.41$\%$ to 0.6$\%$ and 0.97$\%$) for Cross-Eyed in the Close-World and Open-World protocols, respectively, and a reduction of $4.6\%$ (from 3.94$\%$ to 3.76$\%$) in the PolyU database for the Open-World and single biometric case.
\end{abstract}

\begin{keywords}
Biometrics, Deep Representation, Periocular, Transfer Learning, One-Shot Learning
\end{keywords}

\titlepgskip=-15pt

\maketitle

\section{Introduction}
\label{intro}
\PARstart{C}{onvolutional} Neural Networks (CNNs) have become the standard increasingly in applications of Computer Vision and Pattern Recognition.
From object detection \cite{objectdetection021study} \cite{bochkovskiy2020yolov4} to object recognition \cite{efficiendetobjectrecog}, data generation \cite{imggeneration}, image manipulation \cite{imagetransformation}, CNNs dominate the state-of-the-art.
The popularity and success of CNNs largely stem from their ability to learn and extract highly discriminative features, as well as to easily adapt to different applications such as medical data \cite{medicalsegment}, autonomous driving \cite{autonomousdriving}, or, in our case, biometric recognition \cite{feedbackfernandointro}. 

Nonetheless, to achieve good results, CNNs usually require a substantial amount of varied data to allow the network to learn the abstraction of objects  \cite{dataperformancecnns}. Since acquiring such data is often expensive and infeasible, many researchers are working to make CNNs more efficient \cite{tan2019efficientnet}. 
Transfer Learning is one of the most common approaches to tackling data scarcity. It aims to adapt a network trained for a usually more complex task for which much more training data exist to a new target domain. The idea is to take advantage of the feature extraction power of the pre-trained CNN and fine-tune it for the specific task under consideration. One-shot learning is an extreme case of Transfer Learning, where no data is available to train the network for the new target. Instead, a vector of embedding, or deep representation, is extracted from a class-sample image using a pre-trained network for comparison. Then, distance or similarity-based metrics between the deep representations are used to determine if a new image belongs to the same class. Typically, the last layer before the classification stage is used to extract such deep representations. However, as this paper and previous preliminary studies on periocular recognition show \cite{hernandez2018periocular} \cite{alonso2022cross}, selecting the final layer of the network may not always be the best option. Moreover, as we also study here, the best layer selection depends heavily on the input data normalization as well as the amount and variety of data available when training the model.

With the need for One-Shot Learning appeared newer approaches like the use of Contrastive-Loss \cite{contrastiveloss} and Triplet-Loss \cite{triplet}. In these types of losses, the network is optimized to extract a vector of embedding that maximizes the inter-class distance and, in the case of Triplet-Loss, also minimizes the intra-class distance up to some margin. The approaches for One-Shot Learning explained in this section can, once the network has been trained on a large dataset, be used directly on other datasets for recognition. For example, to use a VGG-Face \cite{cao2018vggface2} network directly on a target Face dataset. 

The eye region is one of the most discriminative areas of the face \cite{eyediscriminative} \cite{alonso2022periocular}. However, it was not until 2009 when \cite{park2009periocular} first introduced the concept of periocular recognition.
They described this new biometric as using the facial area in the immediate vicinity of the eye to recognize a person's identity.
Besides the iris, a well-established biometric trait \cite{nguyen2022deep}, the eye's shape, texture, and subcomponents, like eyebrows, eyelids, commissures, or skin, provide much information that one can exploit to recognize a person. 
This periocular area has proven to achieve high recognition performance \cite{alonso2016survey}, not only for identities but also for soft-biometric traits like gender, ethnicity, and age \cite{talreja2022attribute}\cite{alonso2021soft}\cite{AgeMLSURFSVM}\cite{genderocular}\cite{ethnicity}, while having fewer acquisition constraints than other ocular modalities like the iris. 
However, despite its potential, large periocular datasets are scarce \cite{scarcefeedback}, leading to limited research in this area.
Nonetheless, due to the recent Covid-19 pandemic and the widespread use of face masks, this region has gained significant attention within the biometric community \cite{sharma2023periocular}. 

 In a previous contribution \cite{hernandez2018periocular}, we evaluated a selection of CNNs for One-shot periocular recognition on the UBIPr database. Later in \cite{hernandez2019cross}, we also analyzed the utility of a pre-trained CNN for few-shot cross-spectral periocular recognition on the IMP database. Here, we evaluate a wider selection of networks and databases. We also analyze the effect of other factors, such as image pre-processing, domain adaptation, and data partition. Our contributions are shown below:

 \begin{itemize}
     \item State-of-the-art (SOA) periocular recognition comparison using One-Shot Learning. We report the performance per layer of six widely used CNN architectures (ResNet101v2, DenseNet121, VGG19, Inceptionv3, MobileNetv2, and Xception), as well as the most widely used hand-crafted features in periocular recognition (LBPH, HOG, SIFT) for three different datasets (IMP, PolyU, Cross-Eyed). 
 
     \item We investigate the effect that Domain Adaptation had on the selected networks' deep representation when we used CNNs trained for the periocular modality for Cross-Dataset recognition. We consider the following scenarios: CNNs pretrained with the ImageNet dataset, ImageNet CNNs fine-tuned for periocular recognition with auxiliary datasets, randomly initialized CNNs, and finally, randomly initialized CNNs trained for periocular recognition with auxiliary datasets.

     \item We examined how the acquisition method and input image preprocessing can affect the performance of Deep Representations and best-performing layers.
     
     \item Finally, we also report the generalizability of the best layer found by showcasing how performance varies when using the best-layer information between datasets and same-dataset partitions on the Open-World (OW) and Close-World (CW) cases. 

 \end{itemize}

The rest of the paper is organized as follows. Section \ref{related} frames our work within the related research area of periocular biometrics. Section \ref{DBMetricsProt} presents the databases and networks used, training strategy, and general methodology. Section \ref{Method} explains the paper's experimental framework. Section \ref{Discuss} shows the results and compares them with the state-of-the-art and related databases used. Finally, in Section \ref{Conclusions}, we present the final conclusions obtained from our research.

\Figure[ht!](topskip=0pt, botskip=0pt, midskip=0pt)[width=0.9\textwidth]{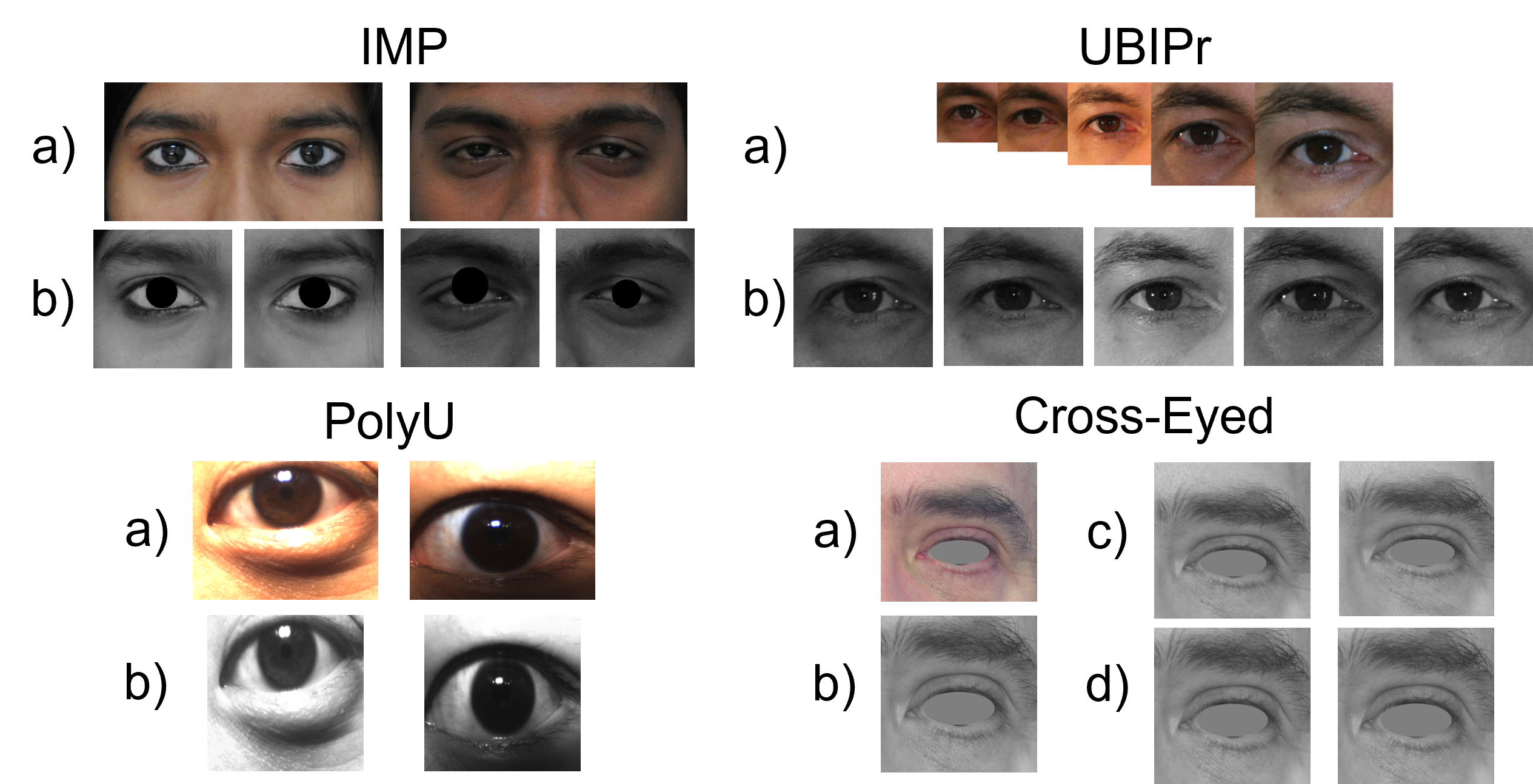}
{Image samples from each database. a) Original database samples, UBIPr shows the relative difference size between image samples b) normalized images after pre-processing as explained in \ref{datasets}. Cross-Eyed c) and d) show the normalization difference used in Section \ref{normalization_effect}. In c), the only modifications to the images were the conversion to grayscale and cropped to be squared, while d) shows the full normalization effect as explained in \ref{datasets}. \label{databases_image}}

\section{Related Work}\label{related}

This section surveys Deep Learning biometric recognition, focusing particularly on periocular biometrics and One-Shot Learning.

This paper extends two previous works \cite{hernandez2018periocular} \cite{hernandez2019cross} that dealt with deep representations for periocular recognition. In \cite{hernandez2018periocular}, we compared the performance per layer of deep representations from a selection of four well-known architectures pre-trained on ImageNet or face recognition databases (AlexNet, GoogLeNet/Inception v1, ResNet, and VGG) and the results with traditional computer vision hand-crafted features. The data employed consisted of periocular images in the visible range from the UBIPr database. This paper discovered that intermediate CNN representations of such networks could outperform traditional CV methods used in biometric recognition with no additional training needed. Furthermore, we saw that biometric-trained models like VGG-Face did not perform better than their general-purpose counterparts trained for the ImageNet challenge. 

In \cite{hernandez2019cross}, we extended the work for cross-spectral periocular recognition using the IMP database, which contains images with three different types of illumination: visible, near-infrared, and night vision. We first analyzed the changes in the performance per layer, observing that the optimal layer is different for each spectrum. We later investigated how intermediate representations could be used for cross-spectral purposes. We found that cross-spectral performance could be improved by training a fully-connected network at the end of the best-performing layer of each spectrum, thus demanding a small fine-tuning step only. The cross-spectral performance was observed to improve largely, up to 65\% (EER) and 87\% (accuracy at 1\% FAR) wrt previous papers, constituting the best-published results to date on the IMP database. The work \cite{hernandez2019cross} focused on one CNN only (ResNet), while the present paper extends the study of the cross-spectral issue to a selection of six different architectures.

The mentioned periocular research of \cite{hernandez2018periocular} \cite{hernandez2019cross} is inspired by \cite{nguyen2017iris}, where the authors studied the per-layer performance in iris recognition of five different ImageNet pre-trained CNNs (AlexNet, VGG, Inception, ResNet, and DenseNet). The segmented and normalized iris image is given to the CNN. The intermediate layers' output at different depths is then extracted and used as feature representations to feed an SVM for identity classification. The paper achieved state-of-the-art recognition performance on two large iris databases, LG2200 (ND-CrossSensor-Iris-2013) and CASIA-Iris-Thousand. The authors concluded that the employed Off-The-Shelf CNNs can extract rich features from iris images that could be used for recognition, thus reducing the complexity of using CNNs for the task by not having to train them, opening the door to new iris representations. In our previous papers \cite{hernandez2018periocular} \cite{hernandez2019cross}, and in the present contribution, we also follow this direction for the periocular modality.

CNNs pretrained on large image datasets such as ImageNet, MS1M, and VGG-Face have been widely used as the backbone of many architectures in the literature. In \cite{talreja2022attribute}, the authors use a frozen VGG16 trained on ImageNet with its Fully Connected layers discarded as the backbone architecture to extract periocular features later used for person recognition, for soft biometric classification, and both together in a Joint Periocular Recognition Block. They improved the SOA for periocular recognition on both UBIRISV2 and FRGC datasets, as well as the soft-biometric classification on FRGC.  

In \cite{surveillance}, the authors used a VGG16 trained for face recognition using the VGG-Face dataset, a dataset with 2,6M images and 2,622 identities, and fine-tuned the network for periocular recognition while controlling the size of the final feature vector. Once the network was adapted to the new domain, the last layers were removed, and the recognition was made by comparing the deep representations of the test images using the Euclidean distance, Spearman distance, or Cosine similarity. They demonstrated the feasibility of using the periocular area in unconstrained scenarios by achieving SOA on NICE.II and MobBIO, two datasets with images captured in uncontrolled environments in the visible spectrum. In another study \cite{impactpreprocessingdeeprepresentation}, the authors used a similar approach to analyze the effect that iris normalization and segmentation have on Deep Representations for biometric recognition. They used two networks (VGG and ResNet50) trained for face recognition and fine-tuned them for iris recognition by removing the last layer and incorporating two new fully connected layers. Once the training was complete, they removed the last classification layer and used the Cosine similarity between the deep representations for biometric verification, reaching a new SOA for the NICE.II dataset.

The authors of \cite{one-shot-triplet} used a One-Shot learning approach to extract a vector of embeddings for joint biometric and sensor recognition. They extracted the images' deep representations using an embedding network that was trained using one of three different types of losses: Cross-Entropy, Contrastive (single and double margin), and Triplet-Loss (with off-line and online triplet mining, as well as multi-class negative-pairs). They then extracted the vector of embedding from the final layer of the network (removing the classification layer from the Cross-entropy approach) and use it for recognition. They compared their results in three different biometric modalities: face, periocular, and iris, as well as for two different types of sensors: Near-Infrared iris sensors and smartphone cameras. They found that the representations were robust across the three biometric modalities and different sensors, outperforming SOA commercial approaches.

In the paper \cite{reddy2020generalizable}, authors proposed a method that consists of a periocular ROI detection model for image alignment, custom data augmentation, and illumination normalization to extract robust and generalizable periocular features using a MobileNetv2 network. They followed an Open-Set protocol in which they trained their models using the VISOB database on visible images and then evaluated the generalizability of their model on UBIRIS-V2, UBIPR, FERET, Cross-Eyed, CASIA-IRIS-TWINS, which includes adverse imaging environment and cross-spectral comparisons, by matching the features vectors extracted using Cosine similarity. They reduced the error rate up to 7 times when compared with existing models in the literature.

In \cite{reddy2019robust}, authors proposed to use an unsupervised convolutional auto-encoder to create subject-invariant feature representations for ocular recognition. For each image input, two augmented views were created and fed to the network. They used an L1 norm between the Deep Representations of both image views created by the encoder and between the original and reconstructed images by the decoder. They coupled the loss also with a KL-divergence term as a sparsity regularizer and two coefficients to weigh the contribution of the regularizer and Deep Representations to the loss. They followed an Open-Set cross-dataset evaluation protocol where they used the Cosine similarity or Hamming distance for matching. They achieved a 2.2\% lower EER for cross-illumination conditions when compared to a supervised ResNet50.

\Figure[h!](topskip=0pt, botskip=0pt, midskip=0pt)[width=0.9\textwidth]{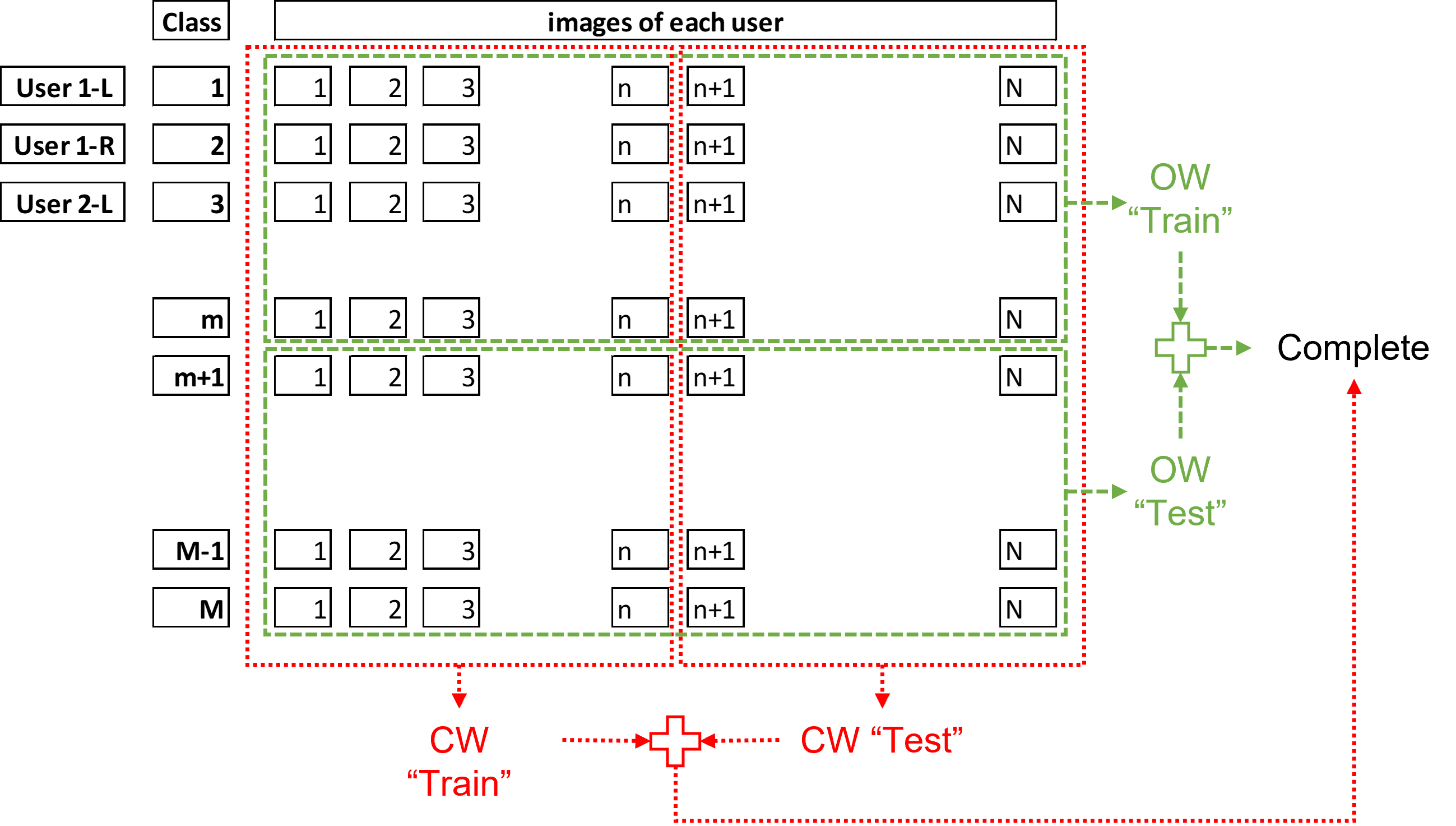}
{Different database partitions for the Close World (CW), Open World (OW), and Complete protocol. \label{data_partitions}}

\begin{table*}[h!]
\centering
\resizebox{0.8\textwidth}{!}{%
\begin{tabular}{|c|c|c|c|c|c|}
\hline
\textbf{Database} & \textbf{Protocol} & \textbf{Train Part. Images (Classes)} & \textbf{Gen./Imp. Pairs} & \textbf{Test Part. Images (Classes)} & \textbf{Gen./Imp. Pairs} \\ \hline

\textbf{PolyU}      & \textbf{CW}       & 4,180 (418) & 18,810/8,715,300 & 2,090 (418) & 4,180/2,178,825  \\ \hline
\textbf{PolyU}      & \textbf{OW}       & 3,135 (209) & 21,945/4,890,600 & 3,135 (209) & 2,1945/4,890,600 \\ \hline
\textbf{Cross-Eyed} & \textbf{Complete} & -           & -                & 1,920 (240) & 6,720/1,835,520  \\ \hline
\textbf{Cross-Eyed} & \textbf{CW}       & 1,200 (240) & 2,400/717,000    & 720 (240)   & 720/258,120      \\ \hline
\textbf{Cross-Eyed} & \textbf{OW}       & 960 (120)   & 3,360/456,960    & 960 (120)   & 3,360/456,960    \\ \hline
\textbf{IMP}        & \textbf{Complete} & - (124)     & -                & 620 (124)   & 1240/190,650     \\ \hline
\end{tabular}
}
\caption{: Summary of Train/Test partitions per database. The Close World (CW) and Open World (OW) protocols with PolyU and Cross-Eyed are defined following \cite{depresion}.}\label{tabla-gen-imp}
\end{table*}

\section{Databases, Metrics, and Protocol}
\label{DBMetricsProt}

This section describes the databases, the matching protocol, and the metrics used to compare the results from our experiments and the baselines.

\subsection{Databases}
\label{datasets}

We employed images in the visible (VIS) range from four commonly used periocular datasets in the experimentation: IIITD Multispectral Periocular (IMP) \cite{IMP}, UBIPr \cite{UBIPr}, Cross-Eyed \cite{x-eyed2016}\cite{x-eyed2017} and PolyU \cite{polyu}.

\textbf{UBIPr} is a periocular database captured with a CANON EOS 5D digital camera with different degrees of subject-camera distance (4-8m), resolutions, illumination, poses, and occlusions in two different sessions. To match the same type of images than the other databases, we only kept the frontal images. In addition, we retained only the users that had two recorded sessions. Since both eyes are available per user per session, our final database has 86 individuals $\times$2 sessions $\times$2 eyes $\times$5 distances = 1720 images. Each eye is considered a different identity, thus having 172 identities. Furthermore, we resized the images using bicubic interpolations. We normalized them (with the annotated ground-truth used in \cite{hernandez2018periocular}) to have the same average sclera radius in their distance group and aligned them by extracting a square region of $7.6R_s$ x $7.6R_s$ around the sclera center.

\textbf{IMP} is a cross-spectral periocular database. It offers images captured in three spectra: Near-Infrared (NIR), Visible (VIS), and Night Vision. The VIS images were captured using a Nikon SLR camera from a distance of 1.3m in a controlled environment and illumination. The database has 62 users with 5 images per user and per spectrum containing both eye regions. We manually annotated the sclera center of each eye and the sclera radius. Then, we separated each eye and normalized the images to have the same sclera radius, and aligned them by cropping a squared region around its sclera center. The database thus has 62 users $\times$2 eyes $\times$5 images per eye = 620 VIS images.

\textbf{Cross-Eyed} is a cross-spectral periocular database captured for the $1^{st}$ Cross-Spectral Iris/Periocular Competition \cite{x-eyed2016}. The database was collected using a custom dual-spectrum image sensor that simultaneously captured images in both NIR and VIS at a distance of 1.5m in an uncontrolled indoor environment. It comprises images of periocular and iris regions of 120 subjects from different nationalities, ethnicities, and eye colors. There are 120 subjects $\times$8 images $\times$2 eyes = 1920 images per spectra and modality. In this paper, we make use of VIS periocular images. Periocular images in Cross-Eyed have their iris masked to ensure pure periocular recognition. We used these masks to normalize them to have the same sclera radius, center, and orientation. They were also zero-padded and cropped, so all have the same size.

\textbf{PolyU} is an iris image database captured using simultaneous bi-spectral imaging. It offers iris images in NIR and VIS, where each eye has pixel correspondence between both spectrum versions. It has 209 subjects $\times$15 images $\times$2 eyes = 6270 images per spectrum. As with the previous datasets, we only used the VIS images for this paper. Since the periocular region in this dataset is rather limited, images are just resized to be squared.

All images were converted to grayscale to normalize skin color across databases, padded with zeroes when images were not squared, resized using bicubic interpolation, and copied across the RGB channels to fit Imagenet networks' input size. Figure \ref{databases_image} shows examples of images from the different databases after this procedure.

\subsection{Metrics}
\label{metrics}

The Equal Error Rate (EER) is the most common evaluation metric for biometric verification systems. EER refers to the error at the intersection point between the False Acceptance Rate (FAR) and the False Rejection Rate (FRR) curves. 

To compare two feature vectors $v1$ and $v2$, we used the cosine similarity illustrated in Equation \ref{cosine} for its fast calculation, even for very high dimensionality vectors, to calculate the FAR and FRR from the CNN embeddings, as well as with the LBPH and HOG descriptors. With SIFT features, we used Equation \ref{sift}, defined as the ratio of matches (M) between images over the minimum number of keypoints (K) detected in either image a and b, with epsilon being a control parameter for any case when no keypoints where found in an image.

\begin{equation}
    \label{cosine}
    cosine = \frac{v_1 * v_2}{\Vert v_1 \Vert \Vert v_2 \Vert}
\end{equation}

\begin{equation}
    \label{sift}
    ratio_{sift} = \frac{M}{\min (K_a,K_b,\epsilon)}
\end{equation}

\subsection{Protocols}

\Figure[hbt!](topskip=0pt, botskip=0pt, midskip=0pt)[width=0.9\textwidth]{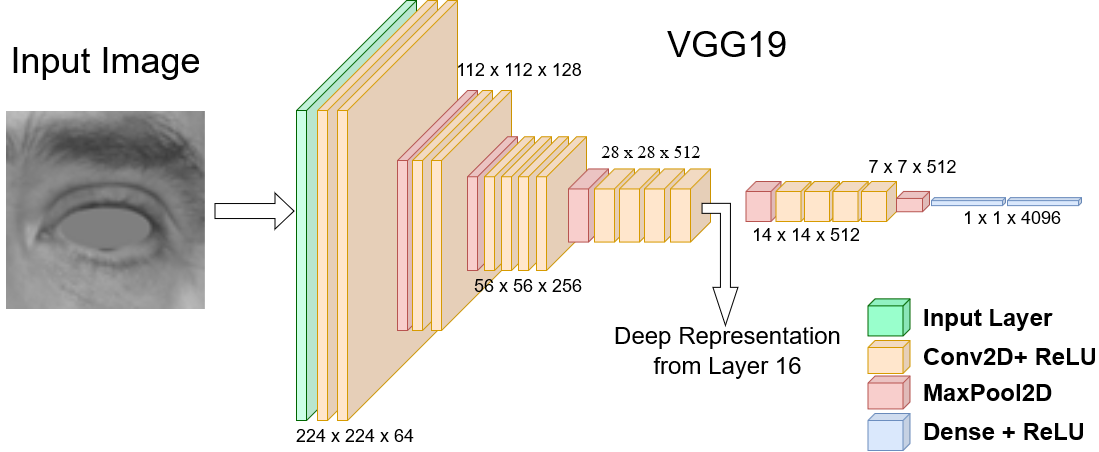}
{Example of a middle layer's Deep Representation extraction for VGG19.\label{middle-layer extraction}}

Although biometric identification was used for training the networks for periocular recognition, a verification setting was the choice for analyzing the performance of the proposed method. In biometric verification, one compares an input image against an image of the identity the user claims to be. If the similarity between images is above a predefined threshold, the user is considered genuine; otherwise, the user is considered an impostor.

All tests employed a cross-dataset One-Shot Learning approach. If a network is to be trained, we used a dataset formed from the combination of all databases introduced in \ref{datasets} except the one used for testing. For instance, when calculating the EER for the Cross-Eyed, a combined dataset with all data from UBIPr, IMP, and PolyU was used to form the training set. Subsequently, for calculating the test performance, we followed an all-against-all strategy, computing all pairs of genuine and impostor scores. All images in the test dataset were used to analyze the performance per layer of the networks. However, we also follow the same protocols as in other previous papers employing the same databases to enable comparison. In particular, for comparison for the PolyU and Cross-Eyed, we use the same approaches carried on \cite{depresion}: the closed-world (CW) protocol, where the images from each user are split into training and testing; and the open-world (OW) protocol, where the users are split into training and test, along with all their available images in such a way that there are no images from the same user in training and test simultaneously. Figure \ref{data_partitions} reflects the difference between the partitions in CW and OW.

In the PolyU dataset and the Close-World setup, the "Test" partition contains the last five images of each user while the remaining ten images are included in the "Train" partition. 
In the Open-World approach, the subjects are divided into two halves of 209 users each, the first half used for the "Train" partition and the latter half of the subjects for the "Test" partition. 
Regarding the Cross-Eyed database, the CW "Test" partition includes the last three images of each user, and the remaining five images go to the "Train" partition. 
For the OW protocol, the users are divided into two halves here as well, with the first 120 users for "Train" and the last 120 users for "Test".

To summarize the number of classes, images, and comparisons for each partition, refer to Table \ref{tabla-gen-imp}. In the case of the IMP database, due to limited data, we only make all-against-all comparisons to calculate the performance on the complete dataset. Finally, the UBIPr dataset is only used in training due to the experiments with different networks already done in \cite{hernandez2018periocular}.

\section{Methodology}\label{Method}

This section presents the experimentation setup used for this study. In particular, the networks, libraries, training strategies, and other algorithms used and how the data was handled and compared.

This paper investigates the performance of deep representations in the middle layers of convolutional neural networks (CNNs) for periocular recognition. We also focus on the impact of training and Transfer Learning on performance. We utilized six widely used and readily available CNNs: ResNet101v2 \cite{resnet}, DenseNet121 \cite{densenet}, VGG19 \cite{vgg}, Xception \cite{xception}, Inceptionv3 \cite{inception}, and MobileNetv2 \cite{mobilenet}. We conducted periocular verification on the VIS images of the IMP, Cross-Eyed, and PolyU datasets presented in the previous section. We assessed how the performance per layer of each network varied as a One-Shot verification algorithm on a target dataset. To do so, four cases were considered: i) networks trained with the ImageNet dataset; ii) ImageNet networks fine-tuned for periocular recognition; iii) random initialized networks; iv) networks trained for periocular recognition from scratch


In cases ii), iv), where the network requires training for periocular recognition, we do so by training it for biometric identification. The training set is composed of all the available periocular datasets except the one used for testing, as indicated in the previous section. When the target (test) dataset was IMP, we combined UBIPr, Cross-Eyed, and PolyU to form a dataset with 830 classes and 9,909 images. We then split it into training and validation sets. The validation set included the last image-distance of each session and user from the UBIPr dataset, the last two images of the Cross-Eyed dataset for each user, and the last five images of each user from PolyU, resulting in training and validation partitions of 6,996 and 2,913 images, respectively. When the target dataset was Cross-Eyed, we trained the networks on a dataset combining UBIPr, IMP, and PolyU, which comprised 8,609 images from 714 classes. The training and validation split followed the same strategy as IMP for UBPIr and PolyU; for the IMP dataset, we used only the last image of each eye and user for the validation split, resulting in training and validation sets of 6,052 and 2,557 images, respectively.

We used Tensorflow-Keras to download, initialize, train, and test the networks. We retained the network's main body, altering only the final Dense layer to fit the number of training classes. We trained them using the Adam optimizer with a learning rate of 0.003, except for VGG, for which we employed Stochastic Gradient Descent with a learning rate of 0.001 and a ClipValue of 0.5, as it provided better stability during training. We trained the networks with Early Stopping, monitoring the validation loss with a patience of 20 epochs and a maximum limit of 500, saving and restoring the weights of the best-performing epoch. Due to GPU memory constraints, the batch sizes were either 16 or 32, depending on the network. We performed data augmentation by randomly rotating the images up to 30 degrees, shifting the height and width by up to 20 percent, and zooming by up to 20 percent. All training was conducted on a Windows 10 machine with 64GB of RAM and an Nvidia RTX2070 GPU with 8GB of VRAM.

\begin{table*}[h!]
\centering
\resizebox{0.8\textwidth}{!}{%
\begin{tabular}{c|cc|cc|cc|cc|}
\cline{2-9}
\textbf{} & \multicolumn{8}{c|}{\textbf{IMP DATABASE}} \\ \cline{2-9}
\multicolumn{1}{l|}{} &
  \multicolumn{8}{c|}{\textbf{TRAINING STRATEGY}} \\ \cline{2-9} 
\textbf{} &
  \multicolumn{2}{c||}{\textbf{i) ImageNet}} &
  \multicolumn{2}{c||}{\textbf{ii) TL ImageNet}} &
  \multicolumn{2}{c||}{\textbf{iii) Random}} &
  \multicolumn{2}{c|}{\textbf{iv) Trained}} \\ \cline{2-9} 
\textbf{} &
  \multicolumn{1}{c|}{\textbf{EER}} &
  \multicolumn{1}{c||}{\textbf{Layer}} &
  \multicolumn{1}{c|}{\textbf{EER}} &
  \multicolumn{1}{c||}{\textbf{Layer}} &
  \multicolumn{1}{c|}{\textbf{EER}} &
  \multicolumn{1}{c||}{\textbf{Layer}} &
  \multicolumn{1}{c|}{\textbf{EER}} &
  \textbf{Layer} \\ \hline
\multicolumn{1}{|c|}{\textbf{ResNet}} &
  \multicolumn{1}{c|}{\cellcolor[HTML]{C0C0C0}\textbf{2,05}} &
  \multicolumn{1}{c||}{218} &
  \multicolumn{1}{c|}{3,31} &
  \multicolumn{1}{c||}{215} &
  \multicolumn{1}{c|}{5,99} &
  \multicolumn{1}{c||}{371} &
  \multicolumn{1}{c|}{\textbf{3,95}} &
  108 \\ \hline
\multicolumn{1}{|c|}{\textbf{DenseNet}} &
  \multicolumn{1}{c|}{3,39} &
  \multicolumn{1}{c||}{101} &
  \multicolumn{1}{c|}{\cellcolor[HTML]{C0C0C0}\textbf{3,26}} &
  \multicolumn{1}{c||}{175} &
  \multicolumn{1}{c|}{5,38} &
  \multicolumn{1}{c||}{438} &
  \multicolumn{1}{c|}{5,37} &
  129 \\ \hline
\multicolumn{1}{|c|}{\textbf{VGG}} &
  \multicolumn{1}{c|}{7,82} &
  \multicolumn{1}{c||}{16} &
  \multicolumn{1}{c|}{5,84} &
  \multicolumn{1}{c||}{16} &
  \multicolumn{1}{c|}{7,33} &
  \multicolumn{1}{c||}{21} &
  \multicolumn{1}{c|}{\cellcolor[HTML]{C0C0C0}4,35} &
  21 \\ \hline
\multicolumn{1}{|c|}{\textbf{Xception}} &
  \multicolumn{1}{c|}{5,07} &
  \multicolumn{1}{c||}{73} &
  \multicolumn{1}{c|}{\cellcolor[HTML]{C0C0C0}4,04} &
  \multicolumn{1}{c||}{73} &
  \multicolumn{1}{c|}{8,26} &
  \multicolumn{1}{c||}{45} &
  \multicolumn{1}{c|}{4,44} &
  53 \\ \hline
\multicolumn{1}{|c|}{\textbf{InceptionV3}} &
  \multicolumn{1}{c|}{\cellcolor[HTML]{C0C0C0}3,79} &
  \multicolumn{1}{c||}{96} &
  \multicolumn{1}{c|}{4,52} &
  \multicolumn{1}{c||}{110} &
  \multicolumn{1}{c|}{5,85} &
  \multicolumn{1}{c||}{163} &
  \multicolumn{1}{c|}{4,23} &
  122 \\ \hline
\multicolumn{1}{|c|}{\textbf{MobileNetV2}} &
  \multicolumn{1}{c|}{6,28} &
  \multicolumn{1}{c||}{96} &
  \multicolumn{1}{c|}{5,75} &
  \multicolumn{1}{c||}{106} &
  \multicolumn{1}{c|}{\cellcolor[HTML]{C0C0C0}\textbf{2,55}} &
  \multicolumn{1}{c||}{157} &
  \multicolumn{1}{c|}{7,91} &
  5 \\ \hline
\end{tabular}
}
\caption{\label{tab:IMP} Best EER performance per network and the layer at which is obtained for the IMP database. The results for all layers are given in Figures \ref{IMP_group} and \ref{Random_group}. The training strategies i)-iv) are detailed in Section \ref{Method}. The numbers in bold indicate the best results per training strategy (colum-wise), while the gray cells indicate the best results per network (row-wise). Results are shown for the Complete protocol defined in Table \ref{tabla-gen-imp}.
}
\end{table*}

After preparing all the networks, we extracted the output of the network layers as illustrated in Figure \ref{middle-layer extraction}. We sliced the network from the input to the desired layer, inputted all the images from the target dataset, extracted the layer's output 4D matrix, and flattened the matrix while maintaining the batch dimension. Subsequently, we compared the entire dataset using an all-against-all matching strategy, employing Cosine similarity for its rapid and straightforward computation before proceeding to the next layer, as described in Section \ref{metrics}. 

We also use in our experiments three methods based on the most widely used features in periocular research, employed as baseline in many studies \cite{alonso2016survey}: Histogram of Oriented Gradients (HOG) \cite{HOG}, Local Binary Patterns (LBPH) \cite{LBP}, and Scale-Invariant Feature Transform (SIFT) key-points \cite{SIFT}. HOG and LBPH features are extracted from non-overlapped regions of the image, forming per-block histograms of 8 bins which are then concatenated to form a feature vector of the entire image. Comparison between two images is done via Cosine similarity between their histograms. On the other hand, SIFT operates by extracting key-points (with dimension 128 per key-point) from the entire image. The comparison metric between two images is as explained in \ref{metrics}. For LBPH and HOG extraction, we used the native Matlab implementation, while for SIFT, we employed the Matlab version available here\footnote{https://www.vlfeat.org/overview/sift.html}.

\section{Results and Discussion}\label{Discuss}

This section presents the results obtained from the experiments for the different networks, databases, and modalities. We started by analyzing the performance of middle-layers representations of well-known networks for periocular verification trained for the ImageNet dataset. These pre-trained networks have become the standard starting point for most image classification tasks \cite{CNN_off_shelf}. Once we obtained the reference results, we explored how they compare when the networks are trained for the same type of data as the target domain, both in EER and depth of the best layer. The results of this study are reported in Section \ref{trainingeffect}. Since we are comparing very high dimensional data using simple similarity scores, we also investigate the impact that alignment and preprocessing on input images can have for this type of Transfer Learning strategy. This is done in Section \ref{normalization_effect}. In the mentioned two sections, we have utilized entire datasets to compare the performance of the methods. To assess how the employed strategies generalize, we examine in Section \ref{section_partition_effect} the consistency of the method in terms of layer depth and performance when changing between training and test partitions on the same dataset as well as at the best layer found on other datasets. Finally, in Section \ref{soacomparison}, we compare our results with previous CNN-based works employing the same datasets, as well as with traditional handcrafted features.

\subsection{Training effect}
\label{trainingeffect}

\Figure[h!](topskip=0pt, botskip=0pt, midskip=0pt)[width=0.9\textwidth]{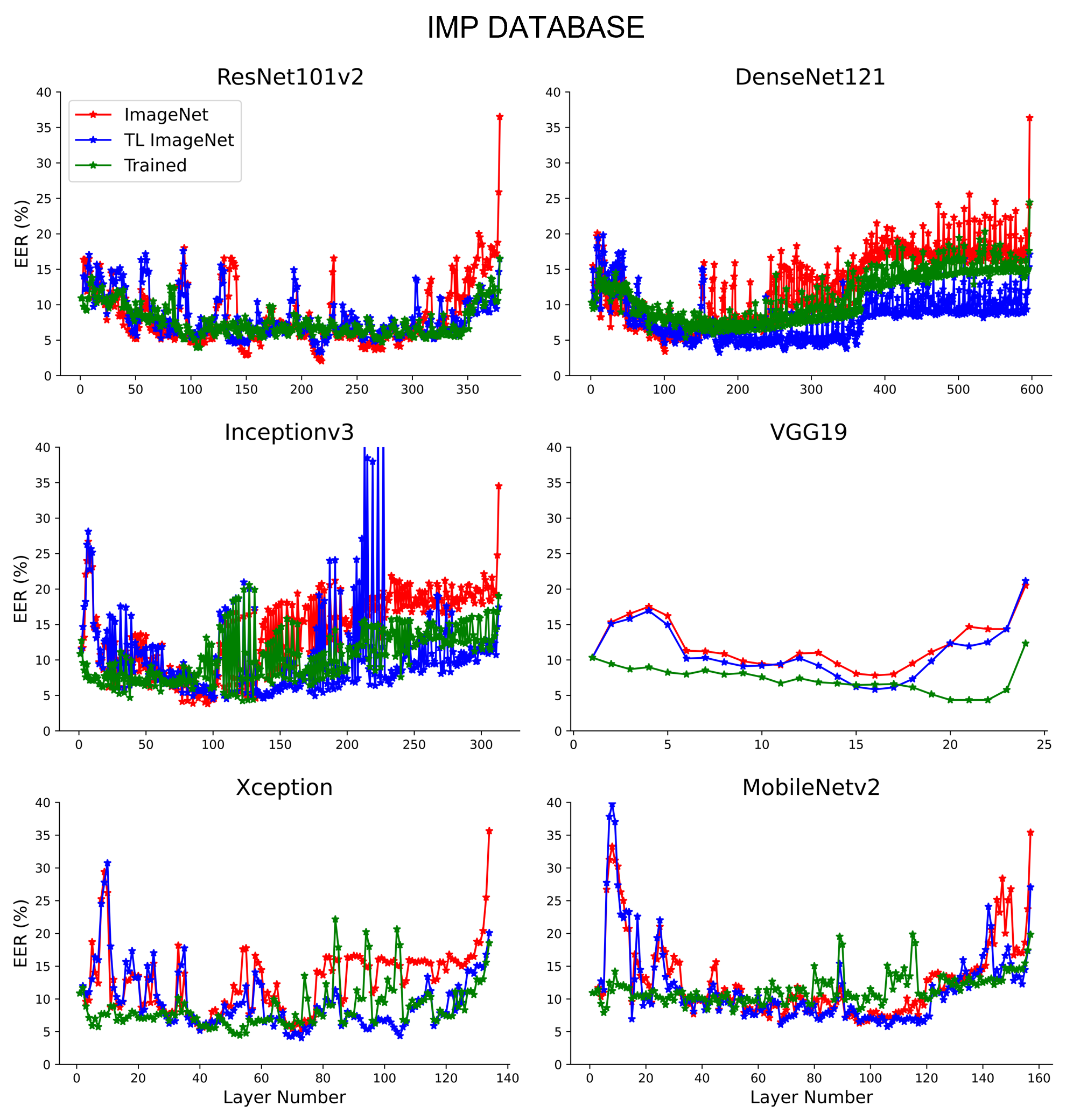}
{EER per layer for the IMP dataset for the cases when: i) the network is trained with the ImageNet dataset (red curve), ii) the network is fine-tuned from ImageNet model for periocular recognition (blue), and iv) the network is trained for periocular recognition from scratch (green curve). These training strategies i), ii), iv) are detailed in Section \ref{Method}. The best cases per network and per training strategy are given in Table \ref{tab:IMP}. Results are shown for the Complete protocol defined in Table \ref{tabla-gen-imp}.\label{IMP_group}}

\Figure[h!](topskip=0pt, botskip=0pt, midskip=0pt)[width=0.9\textwidth]{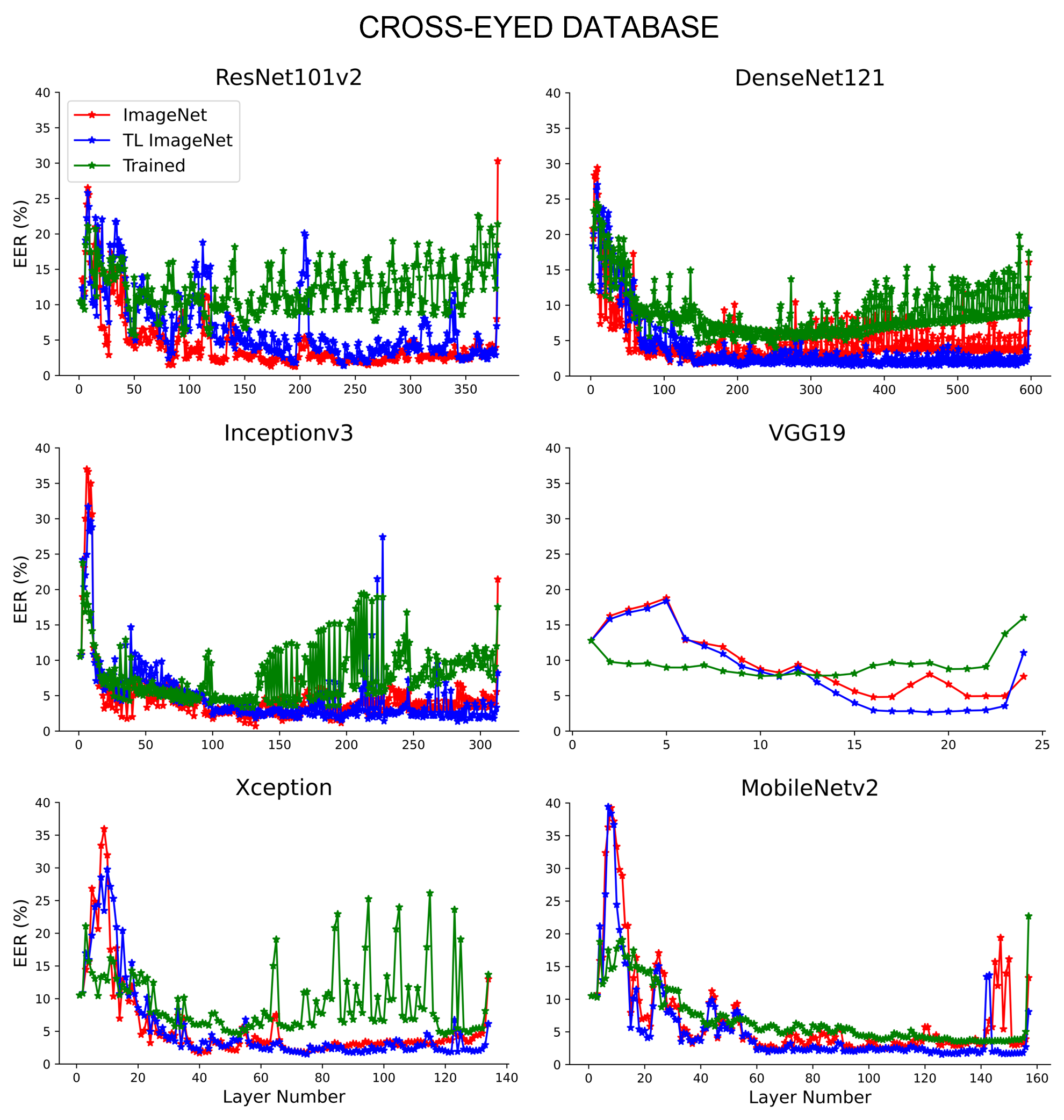}
{EER per layer for the Cross-Eyed dataset for the cases when trained with the ImageNet dataset, when: i) the network is trained with the ImageNet dataset (red curve), ii) the network is fine-tuned from ImageNet model for periocular recognition (blue), and iv) the network is trained for periocular recognition from scratch (green curve). These training strategies i), ii), iv) are detailed in Section \ref{Method}. The best cases per network and per training strategy are given in Table \ref{tab:X-eyed}. Results are shown for the Complete protocol defined in Table \ref{tabla-gen-imp}
\label{XEyed_group}}

\Figure[h!](topskip=0pt, botskip=0pt, midskip=0pt)[width=0.9\textwidth]{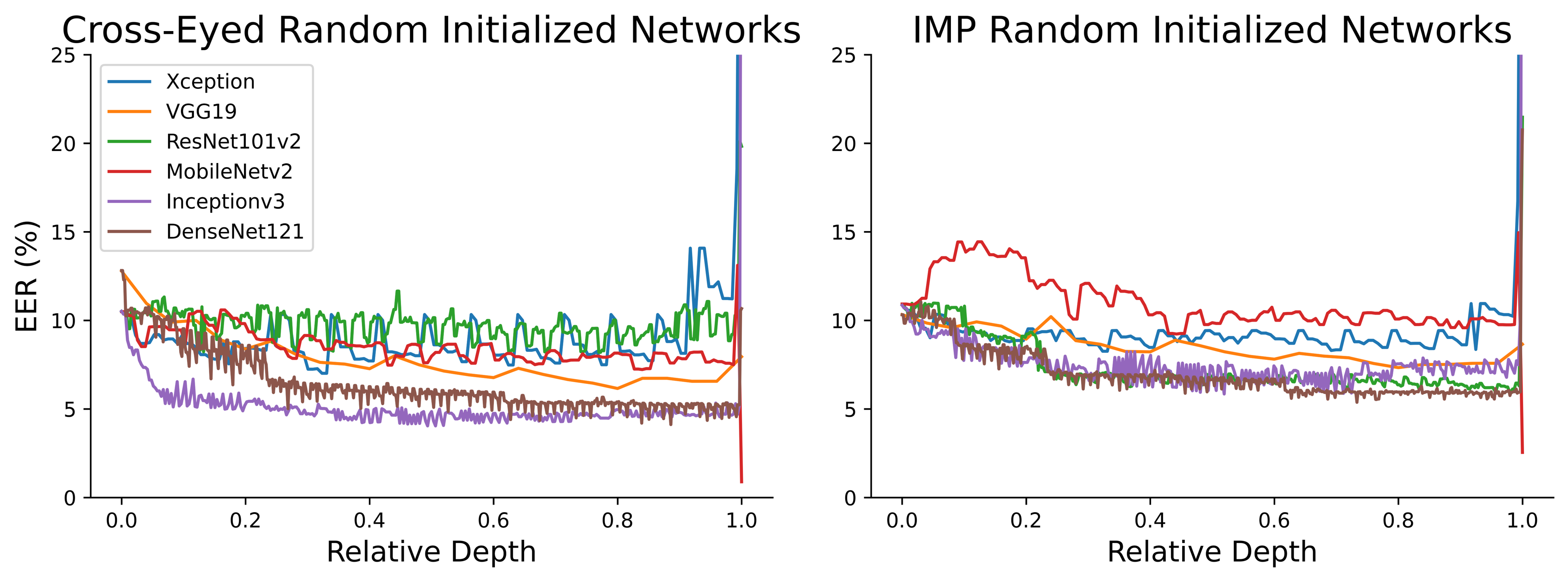}
{EER per layer when the networks were not trained and are just randomly initialized. The best cases per network are given in Table \ref{tab:IMP} (IMP database) and Table \ref{tab:X-eyed} (Cross-Eyed database).  Results are shown for the Complete protocol defined in Table \ref{tabla-gen-imp}.\label{Random_group}}

\begin{table*}[h!]
\centering
\resizebox{0.8\textwidth}{!}{%
\begin{tabular}{c|cc|cc|cc|cc|}
\cline{2-9}
\textbf{} & \multicolumn{8}{c|}{\textbf{CROSS-EYED DATABASE}} \\ \cline{2-9}
\multicolumn{1}{l|}{} &
  \multicolumn{8}{c|}{\textbf{TRAINING STRATEGY}} \\ \cline{2-9} 
\textbf{} &
  \multicolumn{2}{c||}{\textbf{i) ImageNet}} &
  \multicolumn{2}{c||}{\textbf{ii) TL ImageNet}} &
  \multicolumn{2}{c||}{\textbf{iii) Random}} &
  \multicolumn{2}{c|}{\textbf{iv) Trained}} \\ \cline{2-9} 
\textbf{} &
  \multicolumn{1}{c|}{\textbf{EER}} &
  \multicolumn{1}{c||}{\textbf{Layer}} &
  \multicolumn{1}{c|}{\textbf{EER}} &
  \multicolumn{1}{c||}{\textbf{Layer}} &
  \multicolumn{1}{c|}{\textbf{EER}} &
  \multicolumn{1}{c||}{\textbf{Layer}} &
  \multicolumn{1}{c|}{\textbf{EER}} &
  \multicolumn{1}{c|}{\textbf{Layer}} \\ \hline
\multicolumn{1}{|c|}{\textbf{ResNet}} &
  \multicolumn{1}{r|}{\cellcolor[HTML]{C0C0C0}1,25} &
  \multicolumn{1}{r||}{195} &
  \multicolumn{1}{r|}{1,37} &
  \multicolumn{1}{r||}{240} &
  \multicolumn{1}{r|}{8,2} &
  \multicolumn{1}{r||}{292} &
  \multicolumn{1}{r|}{5,65} &
  51 \\ \hline
\multicolumn{1}{|c|}{\textbf{DenseNet}} &
  \multicolumn{1}{r|}{1,67} &
  \multicolumn{1}{r||}{143} &
  \multicolumn{1}{r|}{\cellcolor[HTML]{C0C0C0}1,37} &
  \multicolumn{1}{r||}{462} &
  \multicolumn{1}{r|}{4,12} &
  \multicolumn{1}{r||}{529} &
  \multicolumn{1}{r|}{3,97} &
  252 \\ \hline
\multicolumn{1}{|c|}{\textbf{VGG}} &
  \multicolumn{1}{r|}{4,8} &
  \multicolumn{1}{r||}{16} &
  \multicolumn{1}{r|}{\cellcolor[HTML]{C0C0C0}2,65} &
  \multicolumn{1}{r||}{19} &
  \multicolumn{1}{r|}{6,16} &
  \multicolumn{1}{r||}{21} &
  \multicolumn{1}{r|}{7,76} &
  10 \\ \hline
\multicolumn{1}{|c|}{\textbf{Xception}} &
  \multicolumn{1}{r|}{1,67} &
  \multicolumn{1}{r||}{40} &
  \multicolumn{1}{r|}{\cellcolor[HTML]{C0C0C0}1,49} &
  \multicolumn{1}{r||}{75} &
  \multicolumn{1}{r|}{7,01} &
  \multicolumn{1}{r||}{45} &
  \multicolumn{1}{r|}{4,47} &
  120 \\ \hline
\multicolumn{1}{|c|}{\textbf{InceptionV3}} &
  \multicolumn{1}{r|}{\cellcolor[HTML]{C0C0C0}\textbf{0,7}} &
  \multicolumn{1}{r||}{132} &
  \multicolumn{1}{r|}{\textbf{1,35}} &
  \multicolumn{1}{r||}{283} &
  \multicolumn{1}{r|}{4,04} &
  \multicolumn{1}{r||}{155} &
  \multicolumn{1}{r|}{\textbf{3,31}} &
  130 \\ \hline
\multicolumn{1}{|c|}{\textbf{MobileNetV2}} &
  \multicolumn{1}{r|}{2,11} &
  \multicolumn{1}{r||}{64} &
  \multicolumn{1}{r|}{1,53} &
  \multicolumn{1}{r||}{126} &
  \multicolumn{1}{r|}{\cellcolor[HTML]{C0C0C0}\textbf{0,89}} &
  \multicolumn{1}{r||}{157} &
  \multicolumn{1}{r|}{3,51} &
  132 \\ \hline
\end{tabular}
}
\caption{\label{tab:X-eyed} Best EER performance per network and the layer at which is obtained for the Cross-Eyed database. The results for all layers are given in Figures \ref{XEyed_group} and \ref{Random_group}. The training strategies i)-iv) are detailed in Section \ref{Method}. The numbers in bold indicate the best results per training strategy (colum-wise), while the gray cells indicate the best results per network (row-wise). Results are shown for the Complete protocol defined in Table \ref{tabla-gen-imp}.
}
\end{table*}


Tables \ref{tab:IMP} and \ref{tab:X-eyed}, along with Figures \ref{IMP_group}, \ref{XEyed_group} and \ref{Random_group} show the effect that training has on the deep representation of the networks for periocular verification. using IMP and Cross-Eyed as test databases. The figures show the performance of the different CNN layers per network and per training strategy, while the tables summarize the best performance and in which layer it is obtained. Results in this subsection make use of the "Complete" partition of the databases (Table \ref{tabla-gen-imp}).

As the tables show, the best results are not necessarily obtained with fine-tuned networks (cases ii, iv). For some networks, it is better to use ImageNet weights directly (case i), or even random weights (case iii), as with MobiletNet.. This is consistent with previous findings in \cite{hernandez2018periocular}, in which the VGG-Face network, a VGG network trained for face recognition on a dataset with 1 million images, achieved worse periocular recognition results than its ImageNet counterpart. In absolute numbers, ResNet for the IMP dataset and InceptionV3 for Cross-Eyed are the networks that obtained the best results (EER of 2.05 and 0.7, respectively). ResNet performs the best in the IMP dataset for both the ImageNet (case i in Table \ref{tab:IMP}), and the periocular-trained network (case iv). It also ranks second best for the IMP dataset for the fine-tuned ImageNet network (case ii, Table \ref{tab:IMP}) and the Cross-Eyed dataset ImageNet and fine-tuned ImageNet (cases i and ii in Table \ref{tab:X-eyed}), albeit sharing the position in this last case with DenseNet. Conversely, InceptionV3 performs the best for these three categories in Cross-Eyed. Overall, it is thus unclear what training strategy is optimal since the best EER obtained for each network and dataset varies. For ResNet and InceptionV3, it seems better to use the ImageNet version than any other training. On the other hand, it seems better to fine-tune for DenseNet, VGG, Xception, and MobileNetV2. Interestingly, when the networks are fine-tuned, starting from ImageNet weights ("TL ImageNet" column in the tables or blue curve in the figures) gives better results than starting from scratch consistently ("Trained" column or green curve). This effect is much more prominent in the Cross-Eyed dataset. This corroborates previous works that suggest employing a general purpose training such as ImageNet as starting point, especially if data available for fine-tuning is limited \cite{CNN_off_shelf}.

Upon examining Figures \ref{IMP_group} and \ref{XEyed_group}, we can see that at deeper layers, the fine-tuned networks initialized with ImageNet weights (blue curves) start to perform better than the ImageNet counterpart (red curve). This may be due to the similarity in the domain and higher abstraction at deeper layers achieved by fine-tuning, which helps to close the gap between datasets. However, fine-tuned networks started from scratch (green curves) are, in some cases, even worse than ImageNet networks, especially with the Cross-Eyed database. Again, This confirms that fine-tuning ImageNet networks is a better starting point than scratch initialization, especially with limited training data. On the other hand, at early layers, ImageNet and fine-tuned networks perform very similarly in many cases. This confirms the general assumption that early layers of CNNs usually extract low-level features that are domain-agnostic in many cases, while deeper layers become more specialized for the particular task at hand. Another very relevant result is that the very last layers of the networks always suffer a jump in error, even fine-tuned versions. Moreover, with IMP, many cases show performance degradation even earlier. Indeed, optimal performance with any network or database is attained already at the middle layers or just after the first third of the network. 

We examined the performance per layer of randomly initialized networks as a control (Figure \ref{Random_group}). However, the results are surprisingly good in some cases. As Tables \ref{tab:IMP}, \ref{tab:X-eyed} show, the performance of networks with random weights is not as bad as it could be expected, equating to or even beating other cases involving training. The performance per layer, as seen in Figure \ref{Random_group}, shows a very stable behavior after some initial variability. This is somewhat expected since the weights are random, so the extracted features are also. Most networks have no clear positive or negative tendency as we increase depth, but they show a relative plateau in performance, especially after a relative depth of $0.2$. DenseNet, however, does exhibit a slight performance improvement the deeper the layer is, but this is in the form of steps. Finally, we can see that MobileNet has a peak performance at the very last layer. Indeed, MobileNetv2 with random weights achieves the second-best performance for both datasets. This represents an outlier, but it shows that the exponential behavior of the last layer can also work to one's advantage. Nonetheless, as mentioned above, some randomly initialized networks perform relatively similarly to their trained counterparts. DenseNet exemplifies this for the IMP dataset and InceptionV3 and DenseNet for Cross-Eyed, where the difference in EER is less than $1\%$. A notable example is VGG, which performs better in the random version than its trained one for both datasets. Moreover, as we will analyze later when comparing to other methods (Section \ref{section_partition_effect}, Table \ref{tab:summary}), all randomly initialized networks yield results comparable to baseline CV algorithms like LBPH and HOG.

\subsection{Normalization effect}
\label{normalization_effect}

\Figure[h!](topskip=0pt, botskip=0pt, midskip=0pt)[width=0.9\textwidth]{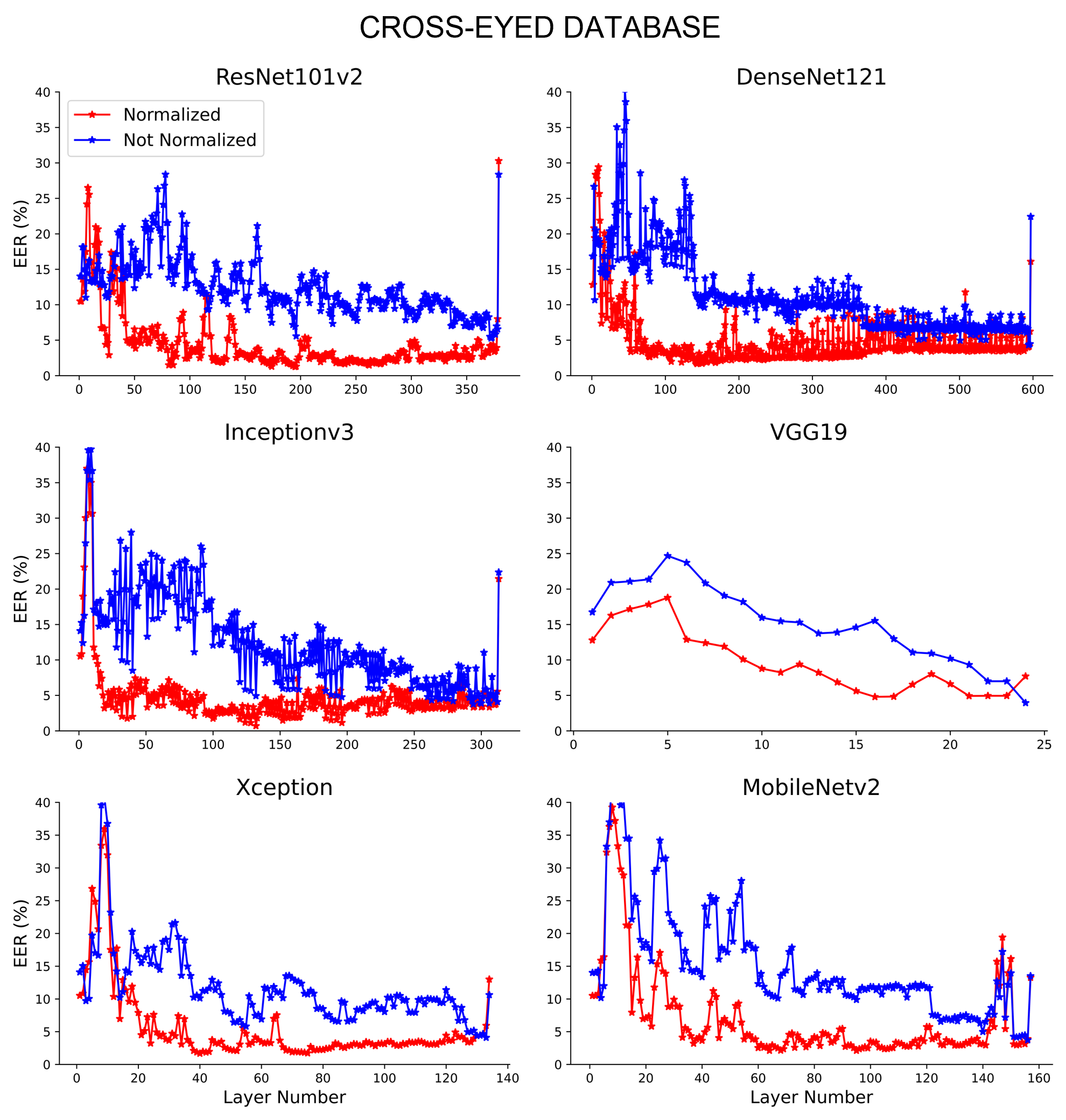}
{EER per layer for the Cross-Eyed dataset for the cases when the images were preprocessed and not. The networks employed are trained with ImageNet. The best cases per network are given in Table \ref{tab:Noprepro}.\label{noprepro_group}}

\begin{table}[h!]
\begin{tabular}{c|cc|cc|c|}
\cline{2-6}
\textbf{} & \multicolumn{5}{c|}{\textbf{CROSS-EYED DATABASE}} \\ \cline{2-6}
\textbf{}                                  & \multicolumn{2}{c||}{\textbf{Normalized}}                        & \multicolumn{2}{c||}{\textbf{Not Normalized}} &              \\ \cline{2-5}
\textbf{} &
  \multicolumn{1}{c|}{\textbf{EER}} &
  \multicolumn{1}{c||}{\textbf{Layer}} &
  \multicolumn{1}{c|}{\textbf{EER}} &
  \multicolumn{1}{c||}{\textbf{Layer}} &
  \multirow{-2}{*}{\textbf{\begin{tabular}[c]{@{}c@{}}Total\\ Layers\end{tabular}}} \\ \hline
\multicolumn{1}{|c|}{\textbf{ResNet}}      & \multicolumn{1}{c|}{\cellcolor[HTML]{C0C0C0}1,25}         & \multicolumn{1}{c||}{195} & \multicolumn{1}{c|}{5,27}           & \multicolumn{1}{c||}{372}  & 379          \\ \hline
\multicolumn{1}{|c|}{\textbf{DenseNet}}    & \multicolumn{1}{c|}{\cellcolor[HTML]{C0C0C0}1,67}         & \multicolumn{1}{c||}{143} & \multicolumn{1}{c|}{4,33}           & \multicolumn{1}{c||}{594}  & 597          \\ \hline
\multicolumn{1}{|c|}{\textbf{VGG}}         & \multicolumn{1}{c|}{4,8}                                  & \multicolumn{1}{c||}{16}  & \multicolumn{1}{c|}{\cellcolor[HTML]{C0C0C0}\textbf{3,68}}  & \multicolumn{1}{c||}{25}   & 26           \\ \hline
\multicolumn{1}{|c|}{\textbf{Xception}}    & \multicolumn{1}{c|}{\cellcolor[HTML]{C0C0C0}1,67}         & \multicolumn{1}{c||}{40}  & \multicolumn{1}{c|}{4,1}            & \multicolumn{1}{c||}{133}  & 134          \\ \hline
\multicolumn{1}{|c|}{\textbf{InceptionV3}} & \multicolumn{1}{c|}{\cellcolor[HTML]{C0C0C0}\textbf{0,7}} & \multicolumn{1}{c||}{132} & \multicolumn{1}{c|}{3,77}           & \multicolumn{1}{c||}{294}  & 313          \\ \hline
\multicolumn{1}{|c|}{\textbf{MobileNetV2}} & \multicolumn{1}{c|}{\cellcolor[HTML]{C0C0C0}2,11}         & \multicolumn{1}{c||}{64}  & \multicolumn{1}{c|}{3,73}           & \multicolumn{1}{c||}{156}  & \textbf{157} \\ \hline
\end{tabular}
\caption{\label{tab:Noprepro} Best EER comparison per network between normalized and not normalized input images and the layer at which is obtained (Cross-Eyed database). The numbers in bold indicate the best results per normalization method (colum-wise), while the gray cells indicate the best results per network (row-wise). The results for all layers are given in Figure \ref{noprepro_group}.}
\end{table}

We then examined the method's robustness for perturbations in the input data and how it affects the deep representation behavior. To do so, we employed normalized and unnormalized images of the Cross-Eyed database. The normalized version consisted of the images processed to have the same sclera radius, center, and orientation, as described in Section \ref{datasets}, whereas the unnormalized images are only converted to grayscale and cropped to be squared using the smallest dimension as reference size. Figure \ref{databases_image} shows the effect of the normalization process on the data for the Cross-Eyed database. Although the Cross-Eyed database was captured at a constant distance, small differences in scale and position between different images can appear if the image is not normalized. Figure \ref{noprepro_group} shows the effect per layer of input normalization on the performance of the deep representations. For space-saving purposes, we report results using networks trained for ImageNet only. We can see that the EER becomes significantly worse if images are unnormalized (blue curves), especially at the early and middle layers. Only in the final layers the performance with unnormalized images becomes closer to the normalized counterparts. This is understandable since networks are susceptible to scale, orientation, and to a minor degree, translation. Only the deeper layers can achieve a higher-level representation of the input data, contributing to closing the gap between the two cases. However, normalized data achieves the best absolute performance with most networks, as shown in Table \ref{tab:Noprepro}. Except for VGG, all networks exhibit an increase in EER between $77\%$ (MobileNetv2) and $439\%$ (InceptionV3) with unnormalized data. We can also see that the best-performing layer with unnormalized data becomes close to the last layer, compared to the normalized version, which usually achieves the best performance in the first half of the network. Thus, robust data normalization is key to achieving better performance.

\subsection{Partition effect}
\label{section_partition_effect}

\begin{table*}[h!]
\centering
\resizebox{0.8\textwidth}{!}{%
\begin{tabular}{cccc|ccc|}
\cline{5-7}
 &
   &
   &
  \textbf{} &
  \multicolumn{3}{c|}{\textbf{Cross-Eyed Test Partition}} \\ \cline{2-7} 
\multicolumn{1}{c|}{\textbf{}} &
  \multicolumn{1}{c|}{\textbf{Network}} &
  \multicolumn{1}{c|}{\textbf{\begin{tabular}[c]{@{}c@{}}Best Layer\\ at\end{tabular}}} &
  \textbf{\begin{tabular}[c]{@{}c@{}}Layer \\ Number\end{tabular}} &
  \multicolumn{1}{c|}{\textbf{\begin{tabular}[c]{@{}c@{}}Train Part \\ EER (\%)\end{tabular}}} &
  \multicolumn{1}{c|}{\textbf{\begin{tabular}[c]{@{}c@{}}Test Part\\ EER (\%)\end{tabular}}} &
  \textbf{\begin{tabular}[c]{@{}c@{}}Complete\\ EER (\%)\end{tabular}} \\ \hline
\multicolumn{1}{|c|}{\multirow{12}{*}{\textbf{CW}}} &
  \multicolumn{1}{c|}{\multirow{4}{*}{\textbf{ResNet101v2}}} &
  \multicolumn{1}{c|}{\textbf{Train Part}} &
  262 &
  \multicolumn{1}{c|}{0,83} &
  \multicolumn{1}{c|}{1,76} &
  1,45 \\ \cline{3-7} 
\multicolumn{1}{|c|}{} &
  \multicolumn{1}{c|}{} &
  \multicolumn{1}{c|}{\textbf{Test Part}} &
  173 &
  \multicolumn{1}{c|}{1,33} &
  \multicolumn{1}{c|}{1,25} &
  1,5 \\ \cline{3-7} 
\multicolumn{1}{|c|}{} &
  \multicolumn{1}{c|}{} &
  \multicolumn{1}{c|}{\textbf{Complete}} &
  195 &
  \multicolumn{1}{c|}{0,92} &
  \multicolumn{1}{c|}{1,53} &
  1,25 \\ \cline{3-7} 
\multicolumn{1}{|c|}{} &
  \multicolumn{1}{c|}{} &
  \multicolumn{1}{c|}{\textbf{IMP}} &
  218 &
  \multicolumn{1}{c|}{1,17} &
  \multicolumn{1}{c|}{2,19} &
  1,94 \\ \hhline{|~|======|}
\multicolumn{1}{|c|}{} &
  \multicolumn{1}{c|}{\multirow{4}{*}{\textbf{DenseNet121}}} &
  \multicolumn{1}{c|}{\textbf{Train Part}} &
  143 &
  \multicolumn{1}{c|}{1,25} &
  \multicolumn{1}{c|}{1,99} &
  1,67 \\ \cline{3-7} 
\multicolumn{1}{|c|}{} &
  \multicolumn{1}{c|}{} &
  \multicolumn{1}{c|}{\textbf{Test Part}} &
  168 &
  \multicolumn{1}{c|}{1,41} &
  \multicolumn{1}{c|}{1,57} &
  1,82 \\ \cline{3-7} 
\multicolumn{1}{|c|}{} &
  \multicolumn{1}{c|}{} &
  \multicolumn{1}{c|}{\textbf{Complete}} &
  143 &
  \multicolumn{1}{c|}{1,25} &
  \multicolumn{1}{c|}{1,99} &
  1,67 \\ \cline{3-7} 
\multicolumn{1}{|c|}{} &
  \multicolumn{1}{c|}{} &
  \multicolumn{1}{c|}{\textbf{IMP}} &
  101 &
  \multicolumn{1}{c|}{2,82} &
  \multicolumn{1}{c|}{3,34} &
  3,21 \\ \hhline{|~|======|}
\multicolumn{1}{|c|}{} &
  \multicolumn{1}{c|}{\multirow{4}{*}{\textbf{Inceptionv3}}} &
  \multicolumn{1}{c|}{\textbf{Train Part}} &
  132 &
  \multicolumn{1}{c|}{0,54} &
  \multicolumn{1}{c|}{0,56} &
  0,7 \\ \cline{3-7} 
\multicolumn{1}{|c|}{} &
  \multicolumn{1}{c|}{} &
  \multicolumn{1}{c|}{\textbf{Test Part}} &
  132 &
  \multicolumn{1}{c|}{0,54} &
  \multicolumn{1}{c|}{0,56} &
  0,7 \\ \cline{3-7} 
\multicolumn{1}{|c|}{} &
  \multicolumn{1}{c|}{} &
  \multicolumn{1}{c|}{\textbf{Complete}} &
  132 &
  \multicolumn{1}{c|}{0,54} &
  \multicolumn{1}{c|}{0,56} &
  0,7 \\ \cline{3-7} 
\multicolumn{1}{|c|}{} &
  \multicolumn{1}{c|}{} &
  \multicolumn{1}{c|}{\textbf{IMP}} &
  96 &
  \multicolumn{1}{c|}{1,89} &
  \multicolumn{1}{c|}{2,59} &
  2,29 \\ \hline\hline
\multicolumn{1}{|c|}{\multirow{12}{*}{\textbf{OW}}} &
  \multicolumn{1}{c|}{\multirow{4}{*}{\textbf{ResNet101v2}}} &
  \multicolumn{1}{c|}{\textbf{Train Part}} &
  196 &
  \multicolumn{1}{c|}{1,07} &
  \multicolumn{1}{c|}{1,45} &
  1,3 \\ \cline{3-7} 
\multicolumn{1}{|c|}{} &
  \multicolumn{1}{c|}{} &
  \multicolumn{1}{c|}{\textbf{Test Part}} &
  195 &
  \multicolumn{1}{c|}{1,19} &
  \multicolumn{1}{c|}{1,2} &
  1,25 \\ \cline{3-7} 
\multicolumn{1}{|c|}{} &
  \multicolumn{1}{c|}{} &
  \multicolumn{1}{c|}{\textbf{Complete}} &
  195 &
  \multicolumn{1}{c|}{1,19} &
  \multicolumn{1}{c|}{1,2} &
  1,25 \\ \cline{3-7} 
\multicolumn{1}{|c|}{} &
  \multicolumn{1}{c|}{} &
  \multicolumn{1}{c|}{\textbf{IMP}} &
  218 &
  \multicolumn{1}{c|}{1,84} &
  \multicolumn{1}{c|}{2,26} &
  1,94 \\ \hhline{|~|======|}
\multicolumn{1}{|c|}{} &
  \multicolumn{1}{c|}{\multirow{4}{*}{\textbf{DenseNet121}}} &
  \multicolumn{1}{c|}{\textbf{Train Part}} &
  143 &
  \multicolumn{1}{c|}{1,58} &
  \multicolumn{1}{c|}{1,73} &
  1,67 \\ \cline{3-7} 
\multicolumn{1}{|c|}{} &
  \multicolumn{1}{c|}{} &
  \multicolumn{1}{c|}{\textbf{Test Part}} &
  150 &
  \multicolumn{1}{c|}{1,72} &
  \multicolumn{1}{c|}{1,7} &
  1,7 \\ \cline{3-7} 
\multicolumn{1}{|c|}{} &
  \multicolumn{1}{c|}{} &
  \multicolumn{1}{c|}{\textbf{Complete}} &
  143 &
  \multicolumn{1}{c|}{2,83} &
  \multicolumn{1}{c|}{3,9} &
  1,67 \\ \cline{3-7} 
\multicolumn{1}{|c|}{} &
  \multicolumn{1}{c|}{} &
  \multicolumn{1}{c|}{\textbf{IMP}} &
  101 &
  \multicolumn{1}{c|}{3,33} &
  \multicolumn{1}{c|}{3,2} &
  3,21 \\ \hhline{|~|======|}
\multicolumn{1}{|c|}{} &
  \multicolumn{1}{c|}{\multirow{4}{*}{\textbf{Inceptionv3}}} &
  \multicolumn{1}{c|}{\textbf{Train Part}} &
  132 &
  \multicolumn{1}{c|}{0,69} &
  \multicolumn{1}{c|}{0,71} &
  0,7 \\ \cline{3-7} 
\multicolumn{1}{|c|}{} &
  \multicolumn{1}{c|}{} &
  \multicolumn{1}{c|}{\textbf{Test Part}} &
  132 &
  \multicolumn{1}{c|}{0,69} &
  \multicolumn{1}{c|}{0,71} &
  0,7 \\ \cline{3-7} 
\multicolumn{1}{|c|}{} &
  \multicolumn{1}{c|}{} &
  \multicolumn{1}{c|}{\textbf{Complete}} &
  132 &
  \multicolumn{1}{c|}{0,69} &
  \multicolumn{1}{c|}{0,71} &
  0,7 \\ \cline{3-7} 
\multicolumn{1}{|c|}{} &
  \multicolumn{1}{c|}{} &
  \multicolumn{1}{c|}{\textbf{IMP}} &
  96 &
  \multicolumn{1}{c|}{2,26} &
  \multicolumn{1}{c|}{2,29} &
  2,29 \\ \hline
\end{tabular}
}
\caption{Table with the best layers and associated EER for each partition (defined in Table \ref{tabla-gen-imp}) of the Cross-Eyed database.}\label{tab:partitionXE}
\end{table*}

\begin{table*}[h!]
\centering
\resizebox{0.8\textwidth}{!}{%
\begin{tabular}{cccc|cc|}
\cline{5-6}
 &
   &
   &
   &
  \multicolumn{2}{c|}{\textbf{PolyU Test Partition}} \\ \cline{2-6} 
\multicolumn{1}{c|}{} &
  \multicolumn{1}{c|}{\textbf{Network}} &
  \multicolumn{1}{c|}{\textbf{\begin{tabular}[c]{@{}c@{}}Best Layer\\ at\end{tabular}}} &
  \textbf{\begin{tabular}[c]{@{}c@{}}Layer \\ Number\end{tabular}} &
  \multicolumn{1}{c|}{\textbf{\begin{tabular}[c]{@{}c@{}}Train Part \\ EER (\%)\end{tabular}}} &
  \textbf{\begin{tabular}[c]{@{}c@{}}Test Part\\ \\ \\ EER (\%)\end{tabular}} \\ \hline
\multicolumn{1}{|c|}{\multirow{12}{*}{\textbf{CW}}} &
  \multicolumn{1}{c|}{\multirow{4}{*}{\textbf{ResNet101v2}}} &
  \multicolumn{1}{c|}{\textbf{Train Part}} &
  373 &
  \multicolumn{1}{c|}{8,25} &
  3,4 \\ \cline{3-6} 
\multicolumn{1}{|c|}{} &
  \multicolumn{1}{c|}{} &
  \multicolumn{1}{c|}{\textbf{Test Part}} &
  373 &
  \multicolumn{1}{c|}{8,25} &
  3,4 \\ \cline{3-6} 
\multicolumn{1}{|c|}{} &
  \multicolumn{1}{c|}{} &
  \multicolumn{1}{c|}{\textbf{\begin{tabular}[c]{@{}c@{}}Cross-Eyed\\ Complete\end{tabular}}} &
  195 &
  \multicolumn{1}{c|}{14,14} &
  7,49 \\ \cline{3-6} 
\multicolumn{1}{|c|}{} &
  \multicolumn{1}{c|}{} &
  \multicolumn{1}{c|}{\textbf{IMP}} &
  218 &
  \multicolumn{1}{c|}{11,23} &
  5,31 \\ \hhline{|~|=====|}
\multicolumn{1}{|c|}{} &
  \multicolumn{1}{c|}{\multirow{4}{*}{\textbf{DenseNet121}}} &
  \multicolumn{1}{c|}{\textbf{Train Part}} &
  594 &
  \multicolumn{1}{c|}{5,79} &
  2,49 \\ \cline{3-6} 
\multicolumn{1}{|c|}{} &
  \multicolumn{1}{c|}{} &
  \multicolumn{1}{c|}{\textbf{Test Part}} &
  594 &
  \multicolumn{1}{c|}{5,79} &
  2,49 \\ \cline{3-6} 
\multicolumn{1}{|c|}{} &
  \multicolumn{1}{c|}{} &
  \multicolumn{1}{c|}{\textbf{\begin{tabular}[c]{@{}c@{}}Cross-Eyed\\ Complete\end{tabular}}} &
  143 &
  \multicolumn{1}{c|}{12,72} &
  6,39 \\ \cline{3-6} 
\multicolumn{1}{|c|}{} &
  \multicolumn{1}{c|}{} &
  \multicolumn{1}{c|}{\textbf{IMP}} &
  101 &
  \multicolumn{1}{c|}{21,08} &
  11,96 \\ \hhline{|~|=====|}
\multicolumn{1}{|c|}{} &
  \multicolumn{1}{c|}{\multirow{4}{*}{\textbf{Inceptionv3}}} &
  \multicolumn{1}{c|}{\textbf{Train Part}} &
  293 &
  \multicolumn{1}{c|}{6,8} &
  2,68 \\ \cline{3-6} 
\multicolumn{1}{|c|}{} &
  \multicolumn{1}{c|}{} &
  \multicolumn{1}{c|}{\textbf{Test Part}} &
  299 &
  \multicolumn{1}{c|}{6,96} &
  2,67 \\ \cline{3-6} 
\multicolumn{1}{|c|}{} &
  \multicolumn{1}{c|}{} &
  \multicolumn{1}{c|}{\textbf{\begin{tabular}[c]{@{}c@{}}Cross-Eyed\\ Complete\end{tabular}}} &
  132 &
  \multicolumn{1}{c|}{8,74} &
  3,91 \\ \cline{3-6} 
\multicolumn{1}{|c|}{} &
  \multicolumn{1}{c|}{} &
  \multicolumn{1}{c|}{\textbf{IMP}} &
  96 &
  \multicolumn{1}{c|}{15,55} &
  8,35 \\ \hline\hline
\multicolumn{1}{|c|}{\multirow{12}{*}{\textbf{OW}}} &
  \multicolumn{1}{c|}{\multirow{4}{*}{\textbf{ResNet101v2}}} &
  \multicolumn{1}{c|}{\textbf{Train Part}} &
  373 &
  \multicolumn{1}{c|}{7,74} &
  7,76 \\ \cline{3-6} 
\multicolumn{1}{|c|}{} &
  \multicolumn{1}{c|}{} &
  \multicolumn{1}{c|}{\textbf{Test Part}} &
  373 &
  \multicolumn{1}{c|}{7,74} &
  7,76 \\ \cline{3-6} 
\multicolumn{1}{|c|}{} &
  \multicolumn{1}{c|}{} &
  \multicolumn{1}{c|}{\textbf{\begin{tabular}[c]{@{}c@{}}Cross-Eyed\\ Complete\end{tabular}}} &
  195 &
  \multicolumn{1}{c|}{13,46} &
  13,54 \\ \cline{3-6} 
\multicolumn{1}{|c|}{} &
  \multicolumn{1}{c|}{} &
  \multicolumn{1}{c|}{\textbf{IMP}} &
  218 &
  \multicolumn{1}{c|}{10,68} &
  10,83 \\ \hhline{|~|=====|}
\multicolumn{1}{|c|}{} &
  \multicolumn{1}{c|}{\multirow{4}{*}{\textbf{DenseNet121}}} &
  \multicolumn{1}{c|}{\textbf{Train Part}} &
  594 &
  \multicolumn{1}{c|}{5,54} &
  5,82 \\ \cline{3-6} 
\multicolumn{1}{|c|}{} &
  \multicolumn{1}{c|}{} &
  \multicolumn{1}{c|}{\textbf{Test Part}} &
  594 &
  \multicolumn{1}{c|}{5,54} &
  5,82 \\ \cline{3-6} 
\multicolumn{1}{|c|}{} &
  \multicolumn{1}{c|}{} &
  \multicolumn{1}{c|}{\textbf{\begin{tabular}[c]{@{}c@{}}Cross-Eyed\\ Complete\end{tabular}}} &
  143 &
  \multicolumn{1}{c|}{12,09} &
  12,34 \\ \cline{3-6} 
\multicolumn{1}{|c|}{} &
  \multicolumn{1}{c|}{} &
  \multicolumn{1}{c|}{\textbf{IMP}} &
  101 &
  \multicolumn{1}{c|}{19,18} &
  21,4 \\ \hhline{|~|=====|} 
\multicolumn{1}{|c|}{} &
  \multicolumn{1}{c|}{\multirow{4}{*}{\textbf{Inceptionv3}}} &
  \multicolumn{1}{c|}{\textbf{Train Part}} &
  294 &
  \multicolumn{1}{c|}{6,56} &
  6,63 \\ \cline{3-6} 
\multicolumn{1}{|c|}{} &
  \multicolumn{1}{c|}{} &
  \multicolumn{1}{c|}{\textbf{Test Part}} &
  279 &
  \multicolumn{1}{c|}{6,92} &
  6,41 \\ \cline{3-6} 
\multicolumn{1}{|c|}{} &
  \multicolumn{1}{c|}{} &
  \multicolumn{1}{c|}{\textbf{\begin{tabular}[c]{@{}c@{}}Cross-Eyed\\ Complete\end{tabular}}} &
  132 &
  \multicolumn{1}{c|}{8,71} &
  8,15 \\ \cline{3-6} 
\multicolumn{1}{|c|}{} &
  \multicolumn{1}{c|}{} &
  \multicolumn{1}{c|}{\textbf{IMP}} &
  96 &
  \multicolumn{1}{c|}{14,41} &
  15,35 \\ \hline
\end{tabular}
}
\caption{Table with the best layers and associated EER for each partition (defined in Table \ref{tabla-gen-imp}) of the PolyU database.}\label{tab:partitionPolyU}
\end{table*}

\Figure[h!](topskip=0pt, botskip=0pt, midskip=0pt)[width=0.9\textwidth]{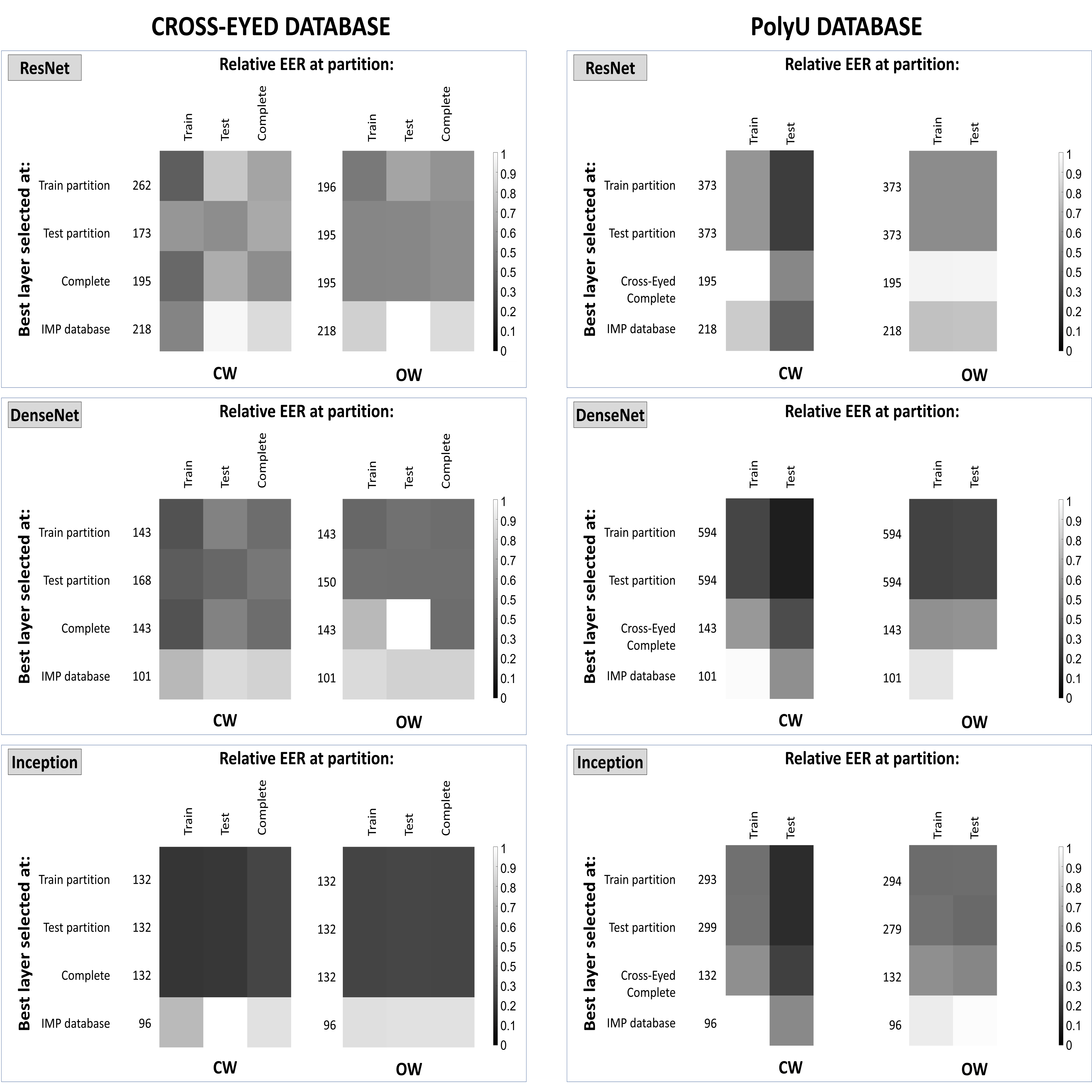}
{Verification accuracy for each partition (defined in Table \ref{tabla-gen-imp}) of Cross-Eyed (left) and PolyU (right). The lower the values, the better. The values are re-scaled, so the maximum EER per CNN and database of the CW/OW experiments are set to 1 (white). Black indicates $0\%$ EER. The exact values are given in Tables \ref{tab:partitionPolyU}, \ref{tab:partitionXE}.\label{cross-partition-performance-image}}

Although one can utilize the network without training them for a specific domain, we have seen in previous subsections that it is still essential to determine which layer yields the best performance. As we demonstrate in Tables \ref{tab:IMP} and \ref{tab:X-eyed}, the best-performing layer can vary significantly from database to database and from network to network, even in the same periocular domain. In the present subsection, we go one step beyond and consider how the performance and the best layer can change considering different partitions of the database. In other words, we select the best layer in one data partition and test the performance in another partition. Furthermore, we check generalizability even further by looking how the performance changes using the best-performing layers from other datasets. For this subsection, we only retain ResNet, DenseNet, and Inception, the best-performing networks of previous subsections (Tables \ref{tab:IMP}, \ref{tab:X-eyed}, and \ref{tab:Noprepro}). We report results using networks trained for ImageNet only for space-saving purposes. In addition, since only PolyU and Cross-Eyed have different partitions and CW/OW protocols (Table \ref{tabla-gen-imp}) \cite{depresion}, this section focuses on these two databases.

Table \ref{tab:partitionXE} shows the EER obtained on Cross-Eyed on the different partitions of the Closed-World (CW), Open-World (OW), and Complete protocols. The "Train", "Test", and "Complete" partitions of the different protocols refer to those detailed in Table \ref{tabla-gen-imp}. Recall that the main difference between CW and OW protocols (Figure \ref{data_partitions}) is that the CW protocol contains the same users in the "Train" and "Test" splits (the images of each user are split into two sets), whereas the OW protocol contains different users in the "Train" and "Test" splits (the users are split into two sets). For better viewing, Figure \ref{cross-partition-performance-image} (left) depicts the relative accuracy values (black=$0\%$, white=maximum EER per CNN/database considering the CW and OW experiments together).

From table \ref{tab:partitionXE} and Figure \ref{cross-partition-performance-image}, Inception offers very stable results with Cross-Eyed, at least when the best layer is selected on a partition of the same database. The best-performing layer for each partition and, when considering the whole dataset, are quite close together, as do the EER (see the relatively similar gray colors in Figure \ref{cross-partition-performance-image}, left bottom). Inception even yields the same exact layer regardless of the partition and mode used to select it. Interestingly, this is true both in the CW case (the Train/Test partitions contain images from the same users) and the OW case (the Train/Test partitions have different users), meaning that Inception generalizes very well over the Cross-Eyed database, even to unseen users. On the other hand, when the best layer is selected externally (using the IMP database), the results degrade substantially, giving the worst EER. Indeed, this is true for any CNN or partition with the Cross-Eyed database (see the brightest boxes in Figure \ref{cross-partition-performance-image}, left, which most of the times correspond to the case when the best layer is selected using IMP). In addition, the best layer with IMP is usually very different than the best layer calculated with other partitions. This suggests that, despite using databases in the VIS spectrum, their differences (in sensor, illumination, etc.) play a very important role in selecting the optimum layer. 

ResNet and Densenet, on the other hand, do not appear to generalize as well as Inception over Cross-Eyed. The best layer is different depending on the partition used to select it, at least in the CW case, where it can also be seen that the performance across partitions varies. This can be appreciated in the gray variations of Figure \ref{cross-partition-performance-image}, left, for the CW case with these two networks. This result is interesting because the CW case contains images from the same users in all partitions, so one would expect similar optimum layers and performance. On the other hand, the OW case contains images from different users on each partition, but the preferred layers are closer, and the performance is more constant across partitions. One observation in this regard would be that the CW protocol entails more users but with fewer images per user in each partition, whereas the OW protocol includes fewer users per partition but with more images per user. This connects with previous research \cite{cao2018vggface2} \cite{alonso2021squeezefaceposenet} that shows that it is better to make training decisions based on a larger number of images per user, even if it implies fewer users. This is because a larger number of samples per user allows to model the intra-class diversity better. In our case, it translates to better generalizability when the best layer is selected under the OW protocol. As seen (Figure \ref{cross-partition-performance-image}, left, OW case), the gray variations between partitions, in this case, are not so high compared to CW.

Table \ref{tab:partitionPolyU} and Figure \ref{cross-partition-performance-image} shows the results for the PolyU dataset. In this case, we can see that the best layers for each network and partition are close to the network's end when they are selected on a partition of the same PolyU database. This is, as explained in Section \ref{normalization_effect}, most likely due to the normalization of the PolyU dataset and its more inconsistent periocular region. Even though several papers use the PolyU dataset as a periocular one, the database was collected to be an iris database, so the surrounding ocular area is not consistent. The periocular area, orientation, location, and scale between images change much more than in Cross-Eyed. Also, the available periocular area is reduced, making periocular recognition with this database challenging. As a result, the layers with the best performance are towards the network's end when the networks have achieved a sufficient level of abstraction. If we use an external database instead, like IMP, which contains periocular images of better quality, the best layers appear earlier for all networks. While the lower abstraction of such layers may be sufficient for IMP, their performance on PolyU is substantially worse.

Regarding the best layer of each network, it can be observed that it is approximately the same when it is selected using PolyU, no matter the partition or protocol used. This is relevant from a generalizability point of view. However, when it comes to the EER, the behavior of the Close-World and Open-World changes drastically, even if the layers at which they are calculated are the same. This can be attributed to the number of comparisons made in each mode combined with the worse quality of the PolyU images. In the Open-World case, since both partitions have the same number of users and the same number of images per user, the EER obtained for each one is very close. However, the number of images per user and partition varies in the Close-World mode, resulting in different genuine and impostor comparisons, as shown in Table \ref{tabla-gen-imp}. As the number of comparisons is smaller in the test partition, especially for the genuine case, the EER on this partition is systematically lower than the train partition. This is because a smaller amount of genuine scores do not allow to account sufficiently for intra-class variability effects, providing a more optimistic performance when the database is of lower quality.

Lastly, it can also be seen with PolyU the negative effect of selecting the optimal layer with an external database. The worst EER (brightest boxes in Figure \ref{cross-partition-performance-image}, right) happens when Cross-Eyed or IMP are used. In addition, there is no consistency per CNN. With ResNet, the worst result is given by the optimal layer in Cross-Eyed. However, with DenseNet and Inception, the worst EER comes from the optimum IMP layers.



\subsection{Comparison with SOA and others}
\label{soacomparison}

\begin{table*}[h!]
\centering
\resizebox{0.8\textwidth}{!}{%
\begin{tabular}{|c|c|c|c|c|c|c|}
\hline
\textbf{Method} &
  \textbf{IMP} &
  \textbf{\begin{tabular}[c]{@{}c@{}}Cross-Eyed\\ Complete\end{tabular}} &
  \textbf{\begin{tabular}[c]{@{}c@{}}Cross-Eyed\\ CW\end{tabular}} &
  \textbf{\begin{tabular}[c]{@{}c@{}}Cross-Eyed\\ OW\end{tabular}} &
  \textbf{\begin{tabular}[c]{@{}c@{}}PolyU\\ CW\end{tabular}} &
  \textbf{\begin{tabular}[c]{@{}c@{}}PolyU\\ OW\end{tabular}} \\ \hline
\textbf{ResNet}                                                & \textbf{2,05} & 1,25         & 1,25          & 1,2           & 3,4           & 7,76          \\ \hline
\textbf{DenseNet}                                              & 3,26          & 1,37         & 1,57          & 1,7           & 2,49          & 5,82          \\ \hline
\textbf{VGG}                                                   & 4,35          & 2,65         & -             & -             & -             & -             \\ \hline
\textbf{Xception}                                              & 4,04          & 1,49         & -             & -             & -             & -             \\ \hline
\textbf{InceptionV3}                                           & 3,79          & \textbf{0,7} & \textbf{0,56} & \textbf{0,71} & 2,67          & 6,41          \\ \hline
\textbf{MobileNetV2}                                           & 2,55          & 0,89         & -             & -             & -             & -             \\ \hline\hline
\textbf{LBPH}                                                  & 6,31          & 3,91         & 2,64          & 5,67          & 7,41          & 13,87         \\ \hline
\textbf{HOG}                                                   & 10,16         & 5,23         & 5,7           & 5,68          & 10,73         & 18,19         \\ \hline
\textbf{SIFT}                                                  & 3,07          & 0,74         & 0,6           & 0,97          & \textbf{1,64} & \textbf{3,76} \\ \hline\hline
\textbf{\begin{tabular}[c]{@{}c@{}}\cite{depresion} \\ Single\end{tabular}} & -             & -            & 1,70          & 3,41          & \textbf{0,78}          & 3,94          \\ \hline
\textbf{\begin{tabular}[c]{@{}c@{}}\cite{depresion} \\ Fusion\end{tabular}} & -             & -            & 1,17          & 2,57          & \textbf{0,35}          & \textbf{2,61}          \\ \hline
\textbf{\cite{vyas2022enhanced}}                                          & -             & -            & 2,41/1,91$^{1}$     & -             & -             & -             \\ \hline
\textbf{\cite{hernandez2019cross}}                                               & 2,95          & -            & -             & -             & -             & -             \\ \hline
\end{tabular}
}
\caption{\label{tab:summary}: Summary of results and comparison with SOA. We reference the best results found in terms of EER for same-spectrum periocular recognition on the visible spectrum for the selected databases. $^{1}$: authors reported the EER for the Left/Right eyes separately}
\end{table*}

Table \ref{tab:summary} summarizes and compares the results with previous works using the same databases \cite{hernandez2019cross}\cite{depresion}\cite{vyas2022enhanced} and the LBP, HOG, and SIFT hand-crafted features. We surpass the state-of-the-art results for the Cross-Eyed dataset achieved by \cite{depresion}. In their study, they train a ResNet50 and a VGG network for cross-spectral periocular recognition but also calculate the performance for same-spectrum verification. InceptionV3 reduced the EER by $58\%$ and $79\%$ for Cross-Eyed in Close-World and Open-World protocols, respectively, and it even achieves superior results than their fusion of iris and periocular. ResNet and DenseNet also reduced the EER w.r.t \cite{depresion} despite lacking training on the target dataset. However, for PolyU, the situation is different. None of the networks achieved comparable or superior results than \cite{depresion}. The increment in the EER for our method is probably due to the higher degree of variability in the PolyU data. At the same time, the bigger gap in the results for the PolyU and Cross-Eyed dataset achieved in \cite{depresion} is partially due to the amount of data available to fine-tune the network, allowing training the CNN better for the task. We also outperformed our previous results on the IMP dataset \cite{hernandez2019cross}. Even if we also used a ResNet101, the version used in this study was a newer ResNet101V2 on Tensorflow-Keras, which can explain the difference.

Regarding traditional computer vision algorithms, LBPH and HOG performed worst on each occasion. Nonetheless, SIFT managed to achieve similar results for all Cross-Eyed partitions. It also achieved comparative results for PolyU in the Close-World protocol but outperformed \cite{depresion} in the Open-World case when the authors only utilized the periocular region. This can be due to the weakness of machine-learning methods when confronted with users that were not present in the training data, which can benefit non-trainable algorithms like SIFT. It must also be highlighted that PolyU has the worst image quality and highest variability among the databases employed. In this case, it becomes less evident the gap between hand-crafted and data-driven approaches when there is limited data.

\section{Conclusions}\label{Conclusions}

This study examined the effect that training and fine-tuning have on the behavior of CNN's deep representations for One-Shot learning. We utilized well-known pretrained networks as out-of-the-box feature extraction methods for periocular recognition. We investigated the behavior per layer of the networks for different datasets and under different training modes. Additionally, we examined the approach's robustness to some natural acquisition noise and how the best layer changes in relation to an auxiliary database or sampled data from the same distribution.

There is no clear best option regarding training strategy. In our experiments, we have observed that it depends on the network used. ResNet, InceptionV3, and MobileNetV2 do better using the ImageNet weights, while the rest can benefit from fine-tuning to the target periocular task. As in previous works, ResNet typically yields one of the best performances among the networks, making it a good default option for this approach. It is worth mentioning that we outperformed CNNs specifically trained not only for the task of biometric recognition but also for the same dataset without having to fine-tune our models. Furthermore, non-CNN-based algorithms like SIFT can still outperform trained CNNs for the same dataset.


Regarding robustness, a crucial factor seems to be the normalization of the input data. Since our method relies on simple similarity scores between high dimensional matrices, misalignment will heavily penalize the performance. Normalization also affects the depth at which the best-performing layer is situated, which tends to be close to the end for not normalized data, which is when the network has achieved a sufficient level of abstraction. When it comes to the sample set used to select the best layer, using an external database with different acquisition conditions has shown to have a very negative effect, giving much worse EER and a very different optimal layer w.r.t. using a partition of the same database. Also, it is essential to have a sufficient number of images per user to properly model intra-user variability. This is especially critical if the target dataset is of very low quality.


A limitation of this approach is that it relies on finding the best layer for the task. To accomplish this, it is necessary to have a certain amount of data to calculate the network's performance for that specific domain properly. Nonetheless, the amount of data needed could be, in principle, smaller than the amount needed to properly train a network, as we can see in Table \ref{tab:summary}, where we achieve better performance for Cross-Eyed than a network trained for biometric recognition with millions of images and then fine-tuned for the dataset. However, the greater the available data, the better results the trained network will have. Another limitation of this approach is that the deep representation matrices of middle layers can be quite big, posing challenges for large datasets or embedded systems due to memory constraints. For these reasons, the normalization process required for this method makes it a suitable option only for small-scale, easy-to-normalize scenarios. Nevertheless, using facial landmarks and iris and sclera segmentation methods, the periocular region is relatively easy to normalize for frontal images. 

Regarding its advantages, our approach can use CNNs as out-of-the-box feature extractors with relatively good results. It also enables us to save resources on training, data collection, and processing power through network pruning, removing all other network parts that are not required to get to the best-performing layer. Our future work on this approach includes investigating methods to reduce the deep representations' dimensionality or memory used, as well as exploring its potential for network pruning in transfer learning.


\section*{Acknowledgment}

The authors thank the Swedish Research Council (VR) and the Swedish Innovation Agency (VINNOVA) for funding their research, as well as the National Supercomputer Center (NSC), funded by Linköping University, for providing the resources necessary for data handling and processing.

\bibliography{refs.bib}

\begin{thebibliography}{10}
\providecommand{\url}[1]{#1}
\csname url@samestyle\endcsname
\providecommand{\newblock}{\relax}
\providecommand{\bibinfo}[2]{#2}
\providecommand{\BIBentrySTDinterwordspacing}{\spaceskip=0pt\relax}
\providecommand{\BIBentryALTinterwordstretchfactor}{4}
\providecommand{\BIBentryALTinterwordspacing}{\spaceskip=\fontdimen2\font plus
\BIBentryALTinterwordstretchfactor\fontdimen3\font minus
  \fontdimen4\font\relax}
\providecommand{\BIBforeignlanguage}[2]{{%
\expandafter\ifx\csname l@#1\endcsname\relax
\typeout{** WARNING: IEEEtran.bst: No hyphenation pattern has been}%
\typeout{** loaded for the language `#1'. Using the pattern for}%
\typeout{** the default language instead.}%
\else
\language=\csname l@#1\endcsname
\fi
#2}}
\providecommand{\BIBdecl}{\relax}
\BIBdecl

\bibitem{objectdetection021study}
E.~Arulprakash and M.~Aruldoss, ``A study on generic object detection with
  emphasis on future research directions,'' \emph{Journal of King Saud
  University-Computer and Information Sciences}, 2021.

\bibitem{bochkovskiy2020yolov4}
A.~Bochkovskiy, C.-Y. Wang, and H.-Y.~M. Liao, ``Yolov4: Optimal speed and
  accuracy of object detection,'' \emph{arXiv preprint arXiv:2004.10934}, 2020.

\bibitem{efficiendetobjectrecog}
Q.~Xie, M.-T. Luong, E.~Hovy, and Q.~V. Le, ``Self-training with noisy student
  improves imagenet classification,'' in \emph{Proceedings of the IEEE/CVF
  Conference on Computer Vision and Pattern Recognition}, 2020, pp.
  10\,687--10\,698.

\bibitem{imggeneration}
T.~Karras, S.~Laine, M.~Aittala, J.~Hellsten, J.~Lehtinen, and T.~Aila,
  ``Analyzing and improving the image quality of stylegan,'' in
  \emph{Proceedings of the IEEE/CVF Conference on Computer Vision and Pattern
  Recognition}, 2020, pp. 8110--8119.

\bibitem{imagetransformation}
E.~Richardson, Y.~Alaluf, O.~Patashnik, Y.~Nitzan, Y.~Azar, S.~Shapiro, and
  D.~Cohen-Or, ``Encoding in style: a stylegan encoder for image-to-image
  translation,'' in \emph{Proceedings of the IEEE/CVF Conference on Computer
  Vision and Pattern Recognition}, 2021, pp. 2287--2296.

\bibitem{medicalsegment}
D.~Jha, M.~A. Riegler, D.~Johansen, P.~Halvorsen, and H.~D. Johansen,
  ``Doubleu-net: A deep convolutional neural network for medical image
  segmentation,'' in \emph{2020 IEEE 33rd International symposium on
  computer-based medical systems (CBMS)}.\hskip 1em plus 0.5em minus
  0.4em\relax IEEE, 2020, pp. 558--564.

\bibitem{autonomousdriving}
V.~R. Kumar, S.~Yogamani, H.~Rashed, G.~Sitsu, C.~Witt, I.~Leang, S.~Milz, and
  P.~M{\"a}der, ``Omnidet: Surround view cameras based multi-task visual
  perception network for autonomous driving,'' \emph{IEEE Robotics and
  Automation Letters}, vol.~6, no.~2, pp. 2830--2837, 2021.

\bibitem{feedbackfernandointro}
\BIBentryALTinterwordspacing
K.~Sundararajan and D.~L. Woodard, ``Deep learning for biometrics: A survey,''
  \emph{ACM Comput. Surv.}, vol.~51, no.~3, may 2018. [Online]. Available:
  \url{https://doi.org/10.1145/3190618}
\BIBentrySTDinterwordspacing

\bibitem{dataperformancecnns}
C.~Luo, X.~Li, L.~Wang, J.~He, D.~Li, and J.~Zhou, ``How does the data set
  affect cnn-based image classification performance?'' in \emph{2018 5th
  International Conference on Systems and Informatics (ICSAI)}.\hskip 1em plus
  0.5em minus 0.4em\relax IEEE, 2018, pp. 361--366.

\bibitem{tan2019efficientnet}
M.~Tan and Q.~Le, ``Efficientnet: Rethinking model scaling for convolutional
  neural networks,'' in \emph{International Conference on Machine
  Learning}.\hskip 1em plus 0.5em minus 0.4em\relax PMLR, 2019, pp. 6105--6114.

\bibitem{hernandez2018periocular}
K.~Hernandez-Diaz, F.~Alonso-Fernandez, and J.~Bigun, ``Periocular recognition
  using cnn features off-the-shelf,'' in \emph{2018 International conference of
  the biometrics special interest group (BIOSIG)}.\hskip 1em plus 0.5em minus
  0.4em\relax IEEE, 2018, pp. 1--5.

\bibitem{alonso2022cross}
F.~Alonso-Fernandez, K.~B. Raja, R.~Raghavendra, C.~Busch, J.~Bigun,
  R.~Vera-Rodriguez, and J.~Fierrez, ``Cross-sensor periocular biometrics in a
  global pandemic: Comparative benchmark and novel multialgorithmic approach,''
  \emph{Information Fusion}, vol.~83, pp. 110--130, 2022.

\bibitem{contrastiveloss}
P.~Khosla, P.~Teterwak, C.~Wang, A.~Sarna, Y.~Tian, P.~Isola, A.~Maschinot,
  C.~Liu, and D.~Krishnan, ``Supervised contrastive learning,'' \emph{Advances
  in neural information processing systems}, vol.~33, pp. 18\,661--18\,673,
  2020.

\bibitem{triplet}
E.~Hoffer and N.~Ailon, ``Deep metric learning using triplet network,'' in
  \emph{Similarity-Based Pattern Recognition: Third International Workshop,
  SIMBAD 2015, Copenhagen, Denmark, October 12-14, 2015. Proceedings 3}.\hskip
  1em plus 0.5em minus 0.4em\relax Springer, 2015, pp. 84--92.

\bibitem{cao2018vggface2}
Q.~Cao, L.~Shen, W.~Xie, O.~M. Parkhi, and A.~Zisserman, ``Vggface2: A dataset
  for recognising faces across pose and age,'' in \emph{2018 13th IEEE
  international conference on automatic face \& gesture recognition (FG
  2018)}.\hskip 1em plus 0.5em minus 0.4em\relax IEEE, 2018, pp. 67--74.

\bibitem{eyediscriminative}
J.~Royer, C.~Blais, I.~Charbonneau, K.~D{\'e}ry, J.~Tardif, B.~Duchaine,
  F.~Gosselin, and D.~Fiset, ``Greater reliance on the eye region predicts
  better face recognition ability,'' \emph{Cognition}, vol. 181, pp. 12--20,
  2018.

\bibitem{alonso2022periocular}
F.~Alonso-Fernandez, J.~Bigun, J.~Fierrez, N.~Damer, H.~Proen{\c{c}}a, and
  A.~Ross, ``Periocular biometrics: A modality for unconstrained scenarios,''
  \emph{arXiv preprint arXiv:2212.13792}, 2022.

\bibitem{park2009periocular}
U.~Park, A.~Ross, and A.~K. Jain, ``Periocular biometrics in the visible
  spectrum: A feasibility study,'' in \emph{2009 IEEE 3rd international
  conference on biometrics: theory, applications, and systems}.\hskip 1em plus
  0.5em minus 0.4em\relax IEEE, 2009, pp. 1--6.

\bibitem{nguyen2022deep}
K.~Nguyen, H.~Proen{\c{c}}a, and F.~Alonso-Fernandez, ``Deep learning for iris
  recognition: A survey,'' \emph{arXiv preprint arXiv:2210.05866}, 2022.

\bibitem{alonso2016survey}
F.~Alonso-Fernandez and J.~Bigun, ``A survey on periocular biometrics
  research,'' \emph{Pattern Recognition Letters}, vol.~82, pp. 92--105, 2016.

\bibitem{talreja2022attribute}
V.~Talreja, N.~M. Nasrabadi, and M.~C. Valenti, ``Attribute-based deep
  periocular recognition: leveraging soft biometrics to improve periocular
  recognition,'' in \emph{Proceedings of the IEEE/CVF Winter Conference on
  Applications of Computer Vision}, 2022, pp. 4041--4050.

\bibitem{alonso2021soft}
F.~Alonso-Fernandez, K.~Hernandez-Diaz, S.~Ramis, F.~J. Perales, and J.~Bigun,
  ``Facial masks and soft-biometrics: Leveraging face recognition cnns for age
  and gender prediction on mobile ocular images,'' \emph{IET Biometrics},
  vol.~10, no.~5, pp. 562--580, 2021.

\bibitem{AgeMLSURFSVM}
K.~K. Kamarajugadda and T.~R. Polipalli, ``Extract features from periocular
  region to identify the age using machine learning algorithms,'' \emph{Journal
  of medical systems}, vol.~43, pp. 1--15, 2019.

\bibitem{genderocular}
A.~Rattani, N.~Reddy, and R.~Derakhshani, ``Convolutional neural networks for
  gender prediction from smartphone-based ocular images,'' \emph{Iet
  Biometrics}, vol.~7, no.~5, pp. 423--430, 2018.

\bibitem{ethnicity}
S.~Khellat-Kihel, J.~Muhammad, Z.~Sun, and M.~Tistarelli, ``Gender and
  ethnicity recognition based on visual attention-driven deep architectures,''
  \emph{Journal of Visual Communication and Image Representation}, vol.~88, p.
  103627, 2022.

\bibitem{scarcefeedback}
L.~A. Zanlorensi, R.~Laroca, E.~Luz, A.~S. Britto, L.~S. Oliveira, and
  D.~Menotti, ``Ocular recognition databases and competitions: a survey,''
  \emph{Artificial Intelligence Review}, vol.~55, no.~1, pp. 129--180, 2022.

\bibitem{sharma2023periocular}
R.~Sharma and A.~Ross, ``Periocular biometrics and its relevance to partially
  masked faces: A survey,'' \emph{Computer Vision and Image Understanding},
  vol. 226, p. 103583, 2023.

\bibitem{hernandez2019cross}
K.~Hernandez-Diaz, F.~Alonso-Fernandez, and J.~Bigun, ``Cross spectral
  periocular matching using resnet features,'' in \emph{2019 International
  Conference on Biometrics (ICB)}.\hskip 1em plus 0.5em minus 0.4em\relax IEEE,
  2019, pp. 1--7.

\bibitem{nguyen2017iris}
K.~Nguyen, C.~Fookes, A.~Ross, and S.~Sridharan, ``Iris recognition with
  off-the-shelf cnn features: A deep learning perspective,'' \emph{IEEE
  Access}, vol.~6, pp. 18\,848--18\,855, 2017.

\bibitem{surveillance}
E.~Luz, G.~Moreira, L.~A.~Z. Junior, and D.~Menotti, ``Deep periocular
  representation aiming video surveillance,'' \emph{Pattern Recognition
  Letters}, vol. 114, pp. 2--12, 2018.

\bibitem{impactpreprocessingdeeprepresentation}
L.~A. Zanlorensi, E.~Luz, R.~Laroca, A.~S. Britto, L.~S. Oliveira, and
  D.~Menotti, ``The impact of preprocessing on deep representations for iris
  recognition on unconstrained environments,'' in \emph{2018 31st SIBGRAPI
  conference on graphics, patterns and images (SIBGRAPI)}.\hskip 1em plus 0.5em
  minus 0.4em\relax IEEE, 2018, pp. 289--296.

\bibitem{one-shot-triplet}
S.~Banerjee and A.~Ross, ``One-shot representational learning for joint
  biometric and device authentication,'' in \emph{2020 25th International
  Conference on Pattern Recognition (ICPR)}.\hskip 1em plus 0.5em minus
  0.4em\relax IEEE, 2021, pp. 5988--5995.

\bibitem{reddy2020generalizable}
N.~Reddy, A.~Rattani, and R.~Derakhshani, ``Generalizable deep features for
  ocular biometrics,'' \emph{Image and Vision Computing}, vol. 103, p. 103996,
  2020.

\bibitem{reddy2019robust}
------, ``Robust subject-invariant feature learning for ocular biometrics in
  visible spectrum,'' in \emph{2019 IEEE 10th International Conference on
  Biometrics Theory, Applications and Systems (BTAS)}.\hskip 1em plus 0.5em
  minus 0.4em\relax IEEE, 2019, pp. 1--9.

\bibitem{depresion}
L.~A. Zanlorensi, D.~R. Lucio, A.~d.~S. Britto~Junior, H.~Proen{\c{c}}a, and
  D.~Menotti, ``Deep representations for cross-spectral ocular biometrics,''
  \emph{IET Biometrics}, vol.~9, no.~2, pp. 68--77, 2020.

\bibitem{IMP}
A.~Sharma, S.~Verma, M.~Vatsa, and R.~Singh, ``On cross spectral periocular
  recognition,'' in \emph{2014 IEEE International Conference on Image
  Processing (ICIP)}.\hskip 1em plus 0.5em minus 0.4em\relax IEEE, 2014, pp.
  5007--5011.

\bibitem{UBIPr}
C.~N. Padole and H.~Proenca, ``Periocular recognition: Analysis of performance
  degradation factors,'' in \emph{2012 5th IAPR international conference on
  biometrics (ICB)}.\hskip 1em plus 0.5em minus 0.4em\relax IEEE, 2012, pp.
  439--445.

\bibitem{x-eyed2016}
A.~Sequeira, L.~Chen, P.~Wild, J.~Ferryman, F.~Alonso-Fernandez, K.~B. Raja,
  R.~Raghavendra, C.~Busch, and J.~Bigun, ``Cross-eyed-cross-spectral
  iris/periocular recognition database and competition,'' in \emph{2016
  International Conference of the Biometrics Special Interest Group
  (BIOSIG)}.\hskip 1em plus 0.5em minus 0.4em\relax IEEE, 2016, pp. 1--5.

\bibitem{x-eyed2017}
A.~F. Sequeira, L.~Chen, J.~Ferryman, P.~Wild, F.~Alonso-Fernandez, J.~Bigun,
  K.~B. Raja, R.~Raghavendra, C.~Busch, T.~de~Freitas~Pereira \emph{et~al.},
  ``Cross-eyed 2017: Cross-spectral iris/periocular recognition competition,''
  in \emph{2017 IEEE International Joint Conference on Biometrics
  (IJCB)}.\hskip 1em plus 0.5em minus 0.4em\relax IEEE, 2017, pp. 725--732.

\bibitem{polyu}
P.~R. Nalla and A.~Kumar, ``Toward more accurate iris recognition using
  cross-spectral matching,'' \emph{IEEE transactions on Image processing},
  vol.~26, no.~1, pp. 208--221, 2016.

\bibitem{resnet}
K.~He, X.~Zhang, S.~Ren, and J.~Sun, ``Deep residual learning for image
  recognition,'' in \emph{Proceedings of the IEEE conference on computer vision
  and pattern recognition}, 2016, pp. 770--778.

\bibitem{densenet}
G.~Huang, Z.~Liu, L.~Van Der~Maaten, and K.~Q. Weinberger, ``Densely connected
  convolutional networks,'' in \emph{Proceedings of the IEEE conference on
  computer vision and pattern recognition}, 2017, pp. 4700--4708.

\bibitem{vgg}
K.~Simonyan and A.~Zisserman, ``Very deep convolutional networks for
  large-scale image recognition,'' in \emph{International Conference on
  Learning Representations}, 2015.

\bibitem{xception}
F.~Chollet, ``Xception: Deep learning with depthwise separable convolutions,''
  in \emph{Proceedings of the IEEE conference on computer vision and pattern
  recognition}, 2017, pp. 1251--1258.

\bibitem{inception}
C.~Szegedy, V.~Vanhoucke, S.~Ioffe, J.~Shlens, and Z.~Wojna, ``Rethinking the
  inception architecture for computer vision,'' in \emph{Proceedings of the
  IEEE conference on computer vision and pattern recognition}, 2016, pp.
  2818--2826.

\bibitem{mobilenet}
M.~Sandler, A.~Howard, M.~Zhu, A.~Zhmoginov, and L.-C. Chen, ``Mobilenetv2:
  Inverted residuals and linear bottlenecks,'' in \emph{Proceedings of the IEEE
  conference on computer vision and pattern recognition}, 2018, pp. 4510--4520.

\bibitem{HOG}
N.~Dalal and B.~Triggs, ``Histograms of oriented gradients for human
  detection,'' in \emph{2005 IEEE computer society conference on computer
  vision and pattern recognition (CVPR'05)}, vol.~1.\hskip 1em plus 0.5em minus
  0.4em\relax Ieee, 2005, pp. 886--893.

\bibitem{LBP}
T.~Ojala, M.~Pietikainen, and T.~Maenpaa, ``Multiresolution gray-scale and
  rotation invariant texture classification with local binary patterns,''
  \emph{IEEE Transactions on pattern analysis and machine intelligence},
  vol.~24, no.~7, pp. 971--987, 2002.

\bibitem{SIFT}
D.~G. Lowe, ``Distinctive image features from scale-invariant keypoints,''
  \emph{International journal of computer vision}, vol.~60, pp. 91--110, 2004.

\bibitem{CNN_off_shelf}
A.~Sharif~Razavian, H.~Azizpour, J.~Sullivan, and S.~Carlsson, ``Cnn features
  off-the-shelf: an astounding baseline for recognition,'' in \emph{Proceedings
  of the IEEE conference on computer vision and pattern recognition workshops},
  2014, pp. 806--813.

\bibitem{alonso2021squeezefaceposenet}
F.~Alonso-Fernandez, J.~Barrachina, K.~Hernandez-Diaz, and J.~Bigun,
  ``Squeezefaceposenet: Lightweight face verification across different poses
  for mobile platforms,'' in \emph{Pattern Recognition. ICPR International
  Workshops and Challenges: Virtual Event, January 10-15, 2021, Proceedings,
  Part VIII}.\hskip 1em plus 0.5em minus 0.4em\relax Springer, 2021, pp.
  139--153.

\bibitem{vyas2022enhanced}
R.~Vyas, ``Enhanced near-infrared periocular recognition through collaborative
  rendering of hand crafted and deep features,'' \emph{Multimedia Tools and
  Applications}, vol.~81, no.~7, pp. 9351--9365, 2022.

\end{thebibliography}
\bibliographystyle{IEEEtran}

\begin{IEEEbiography}[{\includegraphics[width=1in,height=1.25in,clip,keepaspectratio]{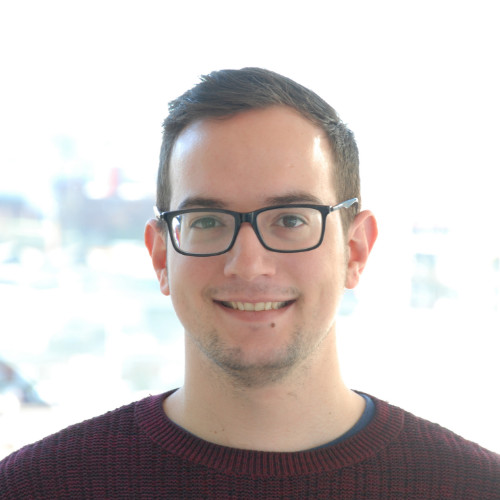}}]{Kevin Hernandez-Diaz}
received the B.S. in telecommunication engineering from Universidad de Sevilla, Spain, in 2016 and the M.S. in Data Science and computer engineering from Universidad de Granada, Spain, in 2017. He is currently pursuing the Ph.D. degree in Signals and Systems Engineering at Halmstad University, Sweden. His thesis focuses on biometric recognition from the ocular region in unconstraint sensing environments. His research interests include AI for biometrics, particularly face and ocular recognition, as well as signal and image analysis, processing, generation, and feature extraction using Deep Learning.
\end{IEEEbiography}

\begin{IEEEbiography}[{\includegraphics[width=1in,height=1.25in,clip,keepaspectratio]{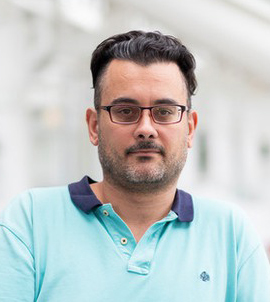}}]{Fernando Alonso-Fernandez} is a docent and an Associate Professor at Halmstad University, Sweden. He received the M.S./Ph.D. degrees in telecommunications from  Universidad Politecnica de Madrid, Spain, in 2003/2008. Since 2010, he is with Halmstad University, Sweden, first as recipient of a Marie Curie IEF and a Postdoctoral Fellowship of the Swedish Research Council, and later with a Project Research Grant for Junior Researchers of the Swedish Research Council. Since 2017, he is Associate Professor at Halmstad University. 

His research interests include AI for biometrics and security, signal and image processing, feature extraction, pattern recognition, and computer vision. He has been involved in multiple EU/national projects focused on biometrics and human–machine interaction. He has over 100 international contributions at refereed conferences and journals and several book chapters. 

Dr. Alonso-Fernandez is Associate Editor of IEEE T-IFS and the IEEE Biometrics Council Newsletter, and an elected member of the IEEE IFS-TC. He co-chaired ICB2016, the 9th IAPR International Conference on Biometrics. He is involved in several roles at the European Association for Biometrics (EAB), such as co-chairing the EAB-RPC or jury member of the of the annual EAB Biometrics Awards.
\end{IEEEbiography}

\begin{IEEEbiography}[{\includegraphics[width=1in,height=1.25in,clip,keepaspectratio]{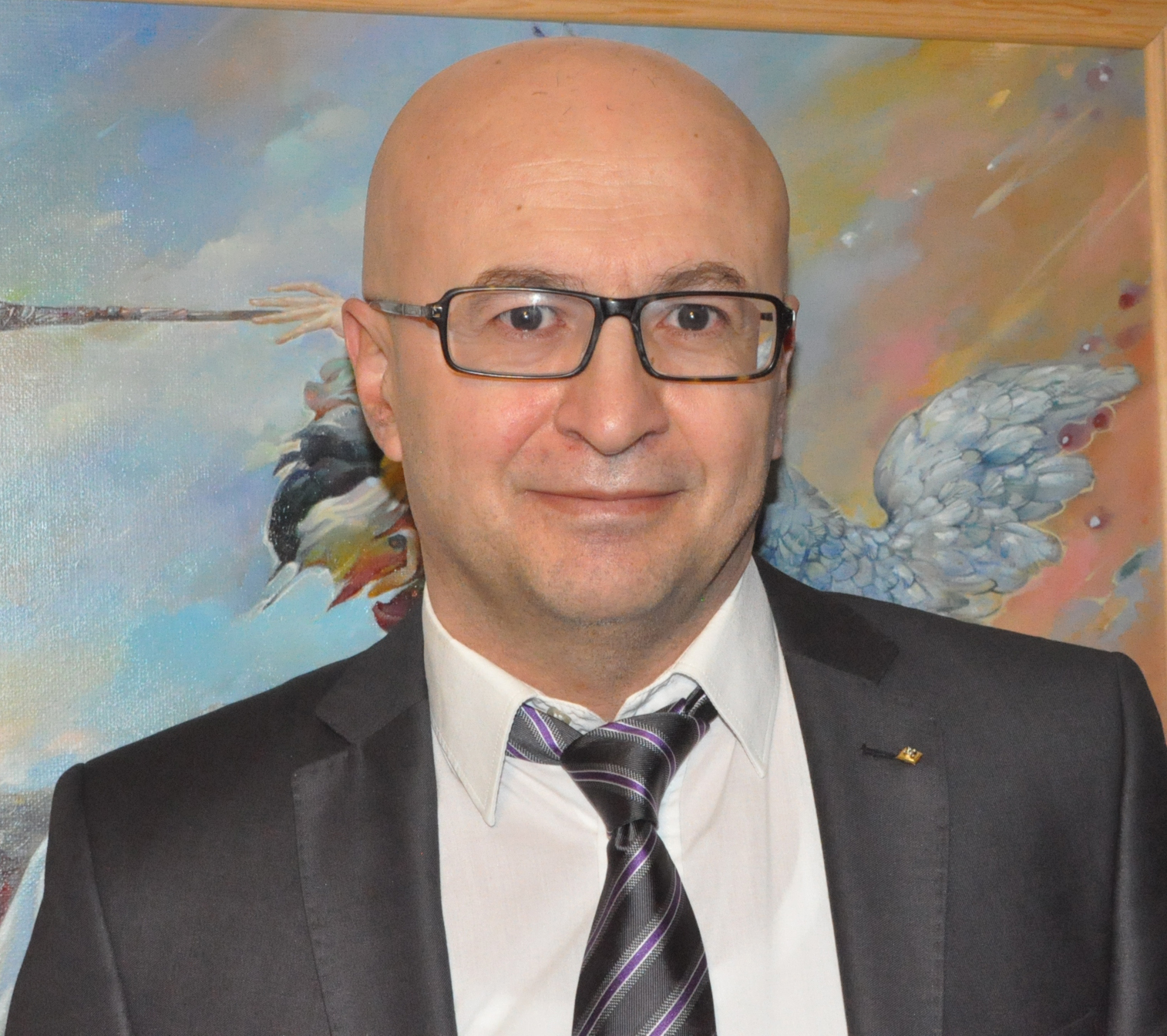}}]{Josef Bigun} is a Full Professor of the Signal Analysis Chair at Halmstad University, Sweden. He received the M.S./Ph.D. degrees from Linköping University, Sweden, in 1983/1988. From 1988 to 1998, he was a Faculty Member with EPFL, Switzerland, as an “Adjoint Scientifique.” He was an Elected Professor of the Signal Analysis Chair (current position) at Halmstad University. 

His scientific interests broadly include computer vision, texture and motion analysis, biometrics, and the understanding of biological recognition mechanisms. He has co-chaired several international conferences and contributed to initiating the ongoing International Conference on Biometrics, formerly known as Audio and Video-Based Biometric Person Authentication, in 1997. He has contributed as Editorial Board Member of journals, including Pattern Recognition Letters, IEEE Transactions on Image Processing, and Image and Vision Computing. He has been Keynote Speaker at several international conferences on pattern recognition, computer vision, and biometrics, including ICPR. He has served on the executive committees of several associations, including IAPR, and as expert for research evaluations, including Sweden, Norway, and EU countries. He is a Fellow of IAPR.
\end{IEEEbiography}

\EOD

\end{document}